\documentclass[lettersize,journal]{IEEEtran}
\usepackage{amsmath,amsfonts}
\usepackage{algorithmic}
\usepackage{algorithm}
\usepackage{array}
\usepackage[caption=false,font=normalsize,labelfont=sf,textfont=sf]{subfig}
\usepackage{textcomp}
\usepackage{stfloats}
\usepackage{url}
\usepackage{verbatim}
\usepackage{graphicx}
\usepackage{cite}
\usepackage{multirow}
\usepackage{booktabs}
\begin{document}
\title{STIP: A SpatioTemporal Information-Preserving and Perception-Augmented Model for High-Resolution Video Prediction}

%\author{Michael~Shell,~\IEEEmembership{Member,~IEEE,}
%        John~Doe,~\IEEEmembership{Fellow,~OSA,}
%        and~Jane~Doe,~\IEEEmembership{Life~Fellow,~IEEE}% <-this % stops a space
%\thanks{M. Shell was with the Department
%of Electrical and Computer Engineering, Georgia Institute of Technology, Atlanta,
\author{Zheng~Chang,
  Xinfeng~Zhang,~\IEEEmembership{Member,~IEEE,}
  Shanshe~Wang,~\IEEEmembership{Member,~IEEE,}
  Siwei~Ma,~\IEEEmembership{Member,~IEEE,}
  %Yan~Ye,~\IEEEmembership{Senior~Member,~IEEE,}
  and~Wen~Gao,~\IEEEmembership{Fellow,~IEEE}% <-this % stops a space
  \thanks{Corresponding author: Siwei Ma (e-mail: swma@pku.edu.cn).}
  \thanks{Z. Chang is with Institute of Computing Technology, Chinese Academy of Sciences,
    Beijing 100190, China, with the University of Chinese Academy of Sciences, Beijing 100049, China, and also with the National Engineering Laboratory for Video Technology, Peking University, Beijing 100871, China
    (e-mail: changzheng18@mails.ucas.ac.cn).}% <-this % stops a space
  \thanks{X. Zhang is with the School of Computer Science and Technology,
    University of Chinese Academy of Sciences, Beijing 100049, China
    (e-mail: xfzhang@ucas.ac.cn).}% <-this % stops a space
  \thanks{S. Wang, S. Ma, and W. Gao are with the National Engineering Laboratory for Video Technology, Peking University, Beijing 100871, China (email: sswang@pku.edu.cn; swma@pku.edu.cn; wgao@pku.edu.cn).}
  %\thanks{Y. Ye was with InterDigital Communications, Inc., San Diego, CA 92121 USA.
  %She is now with Alibaba Group US, Sunnyvale, CA 94085 USA
  %(e-mail: yan.ye@alibaba-inc.com).}
}
\maketitle

\begin{abstract}
  Although significant achievements have been achieved by recurrent neural network (RNN) based video prediction methods, their performance in datasets with high resolutions is still far from satisfactory because of the information loss problem and the perception-insensitive mean square error (MSE) based loss functions. In this paper, we propose a Spatiotemporal Information-Preserving and Perception-Augmented Model (STIP) to solve the above two problems. To solve the information loss problem, the proposed model aims to preserve the spatiotemporal information for videos during the feature extraction and the state transitions, respectively. Firstly, a Multi-Grained Spatiotemporal Auto-Encoder (MGST-AE) is designed based on the X-Net structure.
  The proposed MGST-AE can help the decoders recall multi-grained information from the encoders in both the temporal and spatial domains. In this way, more spatiotemporal information can be preserved during the feature extraction for high-resolution videos. Secondly, a Spatiotemporal Gated Recurrent Unit (STGRU) is designed based on the standard Gated Recurrent Unit (GRU) structure, which can efficiently preserve spatiotemporal information during the state transitions. The proposed STGRU can achieve more satisfactory performance with a much lower computation load compared with the popular Long Short-Term (LSTM) based predictive memories. Furthermore, to improve the traditional MSE loss functions, a Learned Perceptual Loss (LP-loss) is further designed based on the Generative Adversarial Networks (GANs), which can help obtain a satisfactory trade-off between the objective quality and the perceptual quality. Experimental results show that the proposed STIP can predict videos with more satisfactory visual quality compared with a variety of state-of-the-art methods. Source code has been available at \url{https://github.com/ZhengChang467/STIPHR}.
\end{abstract}

% 需要解决的核心问题：信息损失
% 时空编解码
% recalling 机制
% STGRU
% 感知损失
% GAN

% Note that keywords are not normally used for peer-review papers.

% \fi
\IEEEpeerreviewmaketitle

\section{Introduction}\label{sec:introduction}
\IEEEPARstart{V}{ideo} prediction is a very important component of intelligent decision and video processing systems, such as video coding \cite{kang2014depth,zupancic2016inter}, action prediction \cite{li2020spatio}, autonomous driving \cite{kalbkhani2016adaptive,hu2020probabilistic}, video super-resolution \cite{kim2020dynamic}, video interpolation \cite{chen2019uni,liu2017video}, etc. However, predicting the unknown future is full of challenges due to the complex spatiotemporal dynamics in videos. To obtain satisfactory predictions, both the spatial appearance information and the temporal motion information need to be accurately captured.

Inspired by the great successes achieved by deep learning techniques, recurrent neural network (RNN) based methods \cite{chang2022stam,ranzato2014video,srivastava2015unsupervised,shi2015convolutional,finn2016unsupervised,lotter2017deep,villegas2017decomposing,wang2017predrnn,oliu2018folded,wang2018predrnn++,wang2019eidetic,wang2019memory,yu2020efficient,lin2020self,guen2020disentangling,jin2020exploring,wu2021motionrnn,lee2021video,chang2022strpm} have been widely applied in video representation learning due to their great power in processing sequential data. Among all RNN-based methods, the Long Short-Term Memory (LSTM) \cite{hochreiter1997long} based methods can be the most representative one due to their unique advantages in capturing long-short term dependencies. Although some significant improvements have been made by the above methods for video prediction, the performance in predicting videos with high resolutions ($>512$) is still far from satisfactory. There are mainly two reasons accounting for this.

The first reason is the information loss problem during the feature extraction and the state transitions.
On the one hand, high-resolution video inputs typically need to be encoded to low-dimensional features to save computation resources, during which, lots of visual details have to be abandoned.
%To preserve the visual information during extracting deep features, multiple methods \cite{yu2020efficient,jin2020exploring} have been proposed by improving the architecture of encoders, utilizing additional information to improve the visual quality of the results, and so on. However, the low-dimensional deep features still merely contains the high-level semantic information, which are far from enough for decoders to reconstruct satisfactory visual details.
On the other hand, most predictive memories in recent works are still based on the LSTM structure, where the spatiotemporal states are easily influenced by the hidden state and some useful spatiotemporal information may be inevitably filtered out, restricting the performance in modeling reliable spatiotemporal dynamics in videos. In addition, the LSTM-based memories are usually computation-expensive. Although some methods \cite{ballas2016delving,shi2017deep,finn2016unsupervised} have utilized the Gated Recurrent Units (GRUs) \cite{cho2014learning} to help save the computation resources, only temporal information is considered, and the spatial information is severely missing. An efficient and information-preserving spatiotemporal predictive memory urgently needs to be proposed.

The second reason is the perception-insensitive mean square error (MSE) based loss functions, which only average all possible futures to improve the objective scores, such as (Mean Square Error) MSE, (Peak Signal to Noise Ratio) PSNR, (Structural Similarity) SSIM \cite{wang2004image}, etc., leading to unsatisfactory blurry predictions \cite{mathieu2016deep}. To solve this problem, Variational Auto-Encoders (VAEs) \cite{kingma2013auto,babaeizadeh2018stochastic,denton2018stochastic,franceschi2020stochastic,xu2020video,wu2021greedy} and Generative Adversarial Networks (GANs) \cite{babaeizadeh2018stochastic,lee2018stochastic,kwon2019predicting,chen2020long,luc2020transformation} have been utilized to train the predictive models due to their advantages in generating naturalistic results. However, it is difficult to obtain a satisfactory trade-off between the perceptual quality (naturalistic or not) and the objective quality (MSE, SSIM scores).

To solve the above two problems, we propose a spatiotemporal information-preserving and perception-augmented model (STIP) for high-resolution video prediction. This journal paper has greatly improved our previous work \cite{chang2022strpm} in high-resolution video prediction by solving the information loss problem. Different from our previous work \cite{chang2022strpm}, where the encoders have no interactions with the decoders and the predictive unit is still based on the computation-expensive LSTMs, this paper mainly benefits from the following improvements:
\begin{enumerate}
  \item  A multi-grained spatiotemporal auto-encoder is designed to help preserve multi-grained spatiotemporal information during the feature extraction, which is built on the X-Net architecture \cite{fujii2021x}. In both the temporal and spatial domains, multiple skip-connections are employed to help the decoders directly interact with the encoders in different levels of data spaces. In this way, the decoders can recall multi-grained spatiotemporal information from the decoders and more satisfactory visual details can be predicted.
  \item In addition, we further design a new spatiotemporal predictive memory based on the Gated Recurrent Unit (GRU) using more efficient state transitions, which is denoted as Spatiotemporal GRU (STGRU). The STGRU has abandoned the hidden state and only concentrates on the state transitions between the temporal state and the spatial state, where the spatiotemporal information can be well-preserved during the state transitions. Moreover, only two encoders are needed in STIP because of the fewer kinds of states compared with the predictive unit (Residual Predictive Memory: RPM) proposed in our previous work \cite{chang2022strpm} (three encoders are needed), which can further reduce the computational load. Overall, the proposed STGRU are much computation-cheaper with more satisfactory performance compared with the popular LSTM-based memories.
  \item Extensive experiments have been conducted to evaluate the necessity of the extensions.
\end{enumerate}
%On the one hand, to solve the visual information loss problem during feature extraction, multiple encoders and decoders are employed to independently preserve the spatiotemporal information in both spatial and temporal domains. Besides, direct interactions are novelly designed between encoders and decoders using multiple skip-connections, where multi-level visual information can be recalled back by the decoders to reconstruct better prediction.
%Moreover, to further reduce the computation load of the predictive memories, a spatiotemporal gated recurrent unit (STGRU) is proposed, which can achieve similar performance compared with the popular LSTM-based memories with much lower computation load.

To improve the perceptual quality of the predictions, we use the learned perceptual loss (LP-loss) in our previous work \cite{chang2022strpm}, which is designed based on the Generative Adversarial Networks (GANs) to help achieve a satisfactory trade-off between the objective quality and the perceptual quality.

The rest of this paper is organized as follows: Some related works are discussed in Section \ref{sec:relatework}. The proposed method is described in detail in Section \ref{sec:method}. Section \ref{sec:experiment} shows the experimental results and Section \ref{sec:conclusion} concludes this paper.
\section{Related Work}\label{sec:relatework}
\subsection{Memory-based Predictive Methods}
In recent decades, deep learning techniques have made great achievements in extracting deep features for multimedia data, motivated by which, deep learning based methods have been proposed for video prediction. Among all of these methods, the recurrent neural network (RNN) based methods can be the most powerful ones due to their advantages in modeling sequential data. Ranzato \emph{et al.} \cite{ranzato2014video} proposed a baseline model using RNNs for unsupervised video feature learning. To improve the power in capturing long-short term inter-frame dependencies for RNN-based models, Long Short-Term Memory (LSTM) \cite{hochreiter1997long} was employed in \cite{srivastava2015unsupervised} by Srivastava \emph{et al.} for video prediction, which is denoted as FC-LSTM. And to reduce the computation load and improve the model capacity of local perception for videos, Shi \emph{et al.} \cite{shi2015convolutional} replaced the fully connected layers in FC-LSTM with convolutional layers, and the new model was denoted as ConvLSTM, which has achieved some satisfactory improvements for video prediction. Besides LSTMs, Gated Recurrent Unit (GRU) \cite{cho2014learning} has also been improved with convolutional kernels, denoted as ConvGRU \cite{ballas2016delving}. Shi \emph{et al.} \cite{shi2017deep} further extended ConvGRU with location-variant structure for global recurrent connections and Oliu \emph{et al.} introduced a double-mapping GRU (dGRU) to stratify the representation during spatiotemporal learning.

However, most of the above works only focus on capturing the temporal inter-frame dependencies (motion information), the spatial intra-frame features (appearance information) are rarely discussed. Wang \emph{et al.} \cite{wang2017predrnn} proved that the spatial features and temporal dependencies are equally important for predicting high-quality videos. They designed a spatiotemporal LSTM structure (ST-LSTM) and the new predictive model was denoted as PredRNN. To alleviate the gradient propagation difficulties in deep predictive models, Wang \emph{et al.} \cite{wang2018predrnn++} improved PredRNN with gradient highway unit, and the new model was denoted as PredRNN++. Moreover, to capture more complex spatiotemporal signals, they further proposed E3D-LSTM \cite{wang2019eidetic} and MIM \cite{wang2019memory} by integrating 3D convolutional layers and additional forget memory block to ST-LSTMs, respectively. To improve the ability in capturing global dynamics for videos, Lin \emph{et al.} \cite{lin2020self} introduced the self-attention mechanism into ST-LSTMs to memorize long-range spatial features and Lee \emph{et al.} \cite{lee2021video} introduced memory alignment learning to memorize long-term temporal dependencies.
\begin{figure*}
  \centering
  \includegraphics[width=\textwidth]{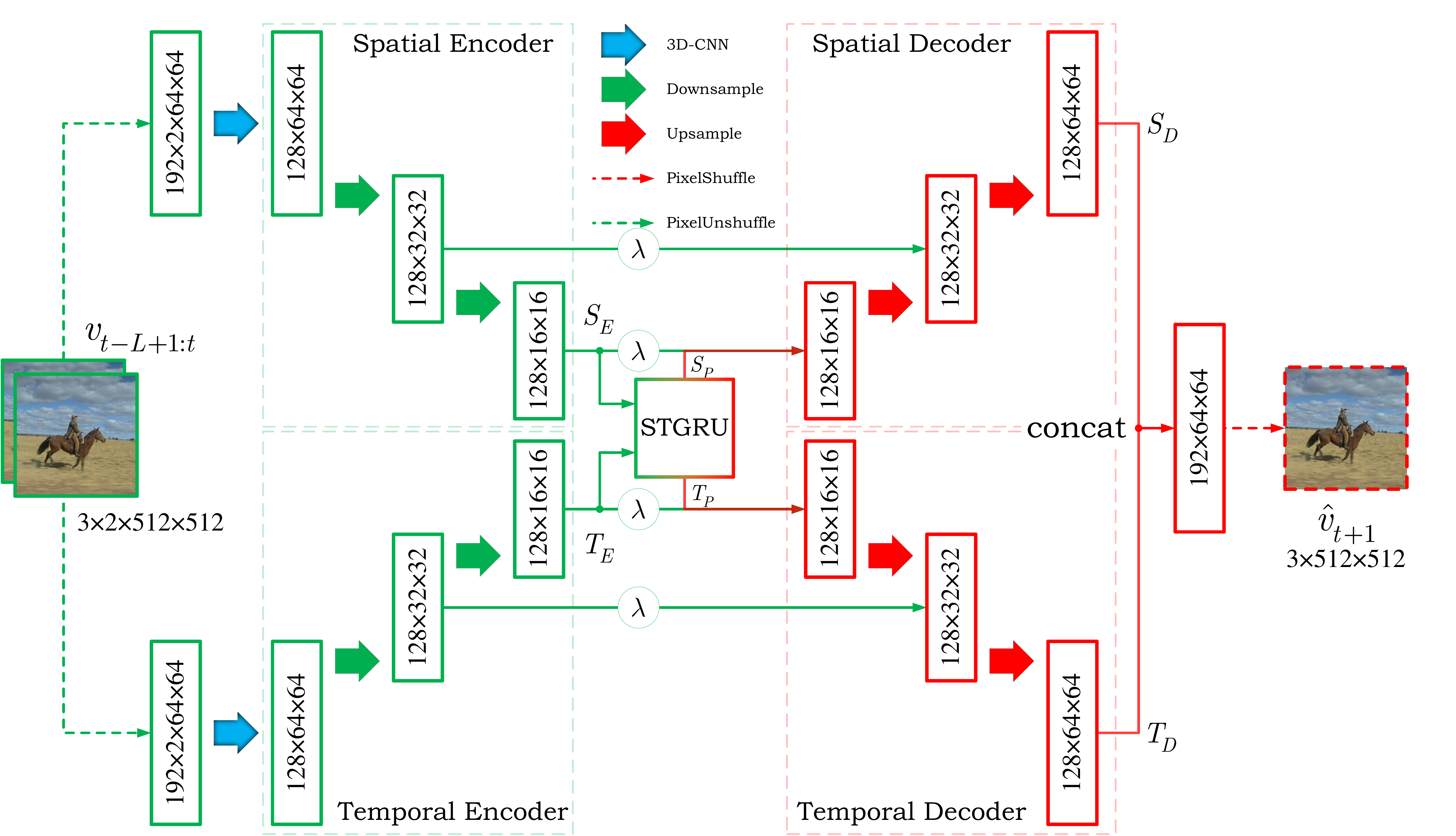}
  \caption{The Multi-Grained Spatiotemporal Auto-encoder: MGST-AE. For each time step, the video input will be encoded to low dimensional features in both the temporal and spatial domains by different encoders. And the decoders will recall multi-grained information from the encoders to reconstruct the visual details of the predictions. Since different features are fed into different modules in STGRU, the STGRU can indirectly supervise different encoders extracting different features in different domains.}\label{fig:system model}
\end{figure*}

\subsection{Structure-based Predictive Methods}
Although some satisfactory results have been obtained by the above works, the datasets used are merely with simple scenarios and low resolutions, such as the Moving MNIST dataset \cite{srivastava2015unsupervised}, KTH action dataset \cite{schuldt2004recognizing}, and so on. The visual details of the predictions on high-resolution real scenarios are severely missing, indicating that merely improving the structure of the predictive memories may not be enough to model more complex videos. 

In recent years, a variety of methods have been proposed to solve this problem via improving the model structures, which can be roughly categorized into two types.
The first type of methods aim to improve the visual quality of the predictions by preserving more visual information during the feature extraction. Yu \emph{et al.} \cite{yu2020efficient} proposed a conditionally reversible network (CrevNet) to preserve the visual information during the feature extraction with a reversible auto-encoder. However, this work has discarded the downsampling operations and the data dimensions of the features are equal to the video inputs, which will greatly increase the computation load for high-resolution video prediction. Jin \emph{et al.} \cite{jin2020exploring} utilized the additional high-frequency information to preserve the details of the predictions. However, the information loss still exists during the frequency information extraction and only limited performance improvements have been obtained. In this paper, we concentrate on solving the information loss problem by proposing a multi-grained spatiotemporal auto-encoder (MGST-AE) and a spatiotemporal gated recurrent unit (STGRU), where the decoders can easily recall multi-grained spatiotemporal information during the whole feature extraction process and the state transitions in the proposed predictive memories are more efficient.

%Although videos with higher PSNR score have been generated in the above works, the blurry details of the predictions are still needed to be enhanced.

The second type of methods aim to improve the visual quality by improving the standard MSE-based loss functions. Mathieu \emph{et al.} \cite{mathieu2016deep} extended the traditional MSE loss function with three different and complementary feature learning strategies. \cite{babaeizadeh2017stochastic,denton2018stochastic,franceschi2020stochastic,xu2020video,wu2021greedy} took advantage of the deep stochastic models to predict different possible futures for different samples based on their latent variables. \cite{lee2018stochastic,liang2017dual,kwon2019predicting,chen2020long,luc2020transformation} utilized GANs to predict videos with more naturalistic appearance. However, the training process in the above methods is very unstable and it is difficult to obtain a satisfactory trade-off between the objective quality and the perceptual quality.
%Although the above works have generated more naturalistic videos, the visual distortions compared with the ground truth are unacceptable (unsatisfactory MSE and SSIM scores) and the fundamental information loss problem during encoding process has still not been fully considered. To focus on the information loss problem in video prediction. The second type of solutions aim to preserve and rich the visual information during feature extraction. Yu \emph{et al.} \cite{yu2020efficient} proposed a conditionally reversible network (CrevNet) to preserve the visual information during feature extraction with a reversible auto-encoder. Jin \emph{et al.} \cite{jin2020exploring} utilized the additional high-frequency information to preserve the details of the predictions. Although videos with higher PSNR score have been generated in the above works, the blurry details of the predictions are still needed to be enhanced.
In our previous work \cite{chang2022strpm}, we designed a learned perceptual loss (LP-loss) to help obtain a more satisfactory trade-off between the objective quality and the perceptual quality.
%on the one hand, we utilize multiple encoders and decoders to process the videos in both temporal and spatial domains, which can preserve more spatiotemporal information during feature extraction. Besides, additional interactions between the corresponding layers of encoders and decoders are applied by using multiple skip-connections, which can help decoders recall more multi-level visual information from encoders for better visual details. On the other hand, besides training the proposed model with GANs for more naturalistic predictions, we also design a learned perceptual loss to further improve the perceptual quality and stabilize the training process. Moreover, a efficient spatiotemporal predictive memory based on GRU (STGRU) is designed, which can achieve similar performance with much lower computational load compared with the popular LSTM-based memories.
% 逼真度入手 GAN ...损失函数入手 Beyond MSE ...
% 增加信息量入手 CrevNet Jin et al.

\section{Method}\label{sec:method}

In this paper, we aim to build an information-preserving and perception-augmented predictive model by solving the following problems,
\begin{itemize}
  \item How to efficiently preserve meaningful spatiotemporal information from the video inputs?
  \item How to predict results with higher perceptual quality?
\end{itemize}
The rest part of Section \ref{sec:method} will elaborate how the proposed model solves the above problems.
% 是否可以通过具体公式表示
\subsection{Building an Information-Preserving Predictive Model}
Based on the analysis in Section \ref{sec:introduction}, the information loss in video prediction tasks mainly comes from the feature extraction and the state transitions. In our previous work \cite{chang2022strpm}, multiple encoders are utilized to extract low-dimensional deep features for different modules of the predictive unit, and the predicted features are decoded back to high-dimensional data using multiple decoders. Although the spatiotemporal information can be preserved during the feature extraction using multiple encoders and decoders, the high-resolution frames are still decoded from merely the low-dimensional features, during which the visual details are hard to be accurately predicted. 
Motivated by the recently published X-Net \cite{fujii2021x}, we design a Multi-Grained Spatiotemporal Auto-Encoder (MGST-AE) to help further preserve the spatiotemporal information during the feature extraction, where decoders can jointly utilize multi-grained features from the encoders to predict more satisfactory visual details. The model structure of MGST-AE is shown in Fig. \ref{fig:system model}. 
% In the improved method, two predictive structures are designed to deal with the information loss problem.

\begin{figure*}[t]
  \centering
  \includegraphics[width=\textwidth]{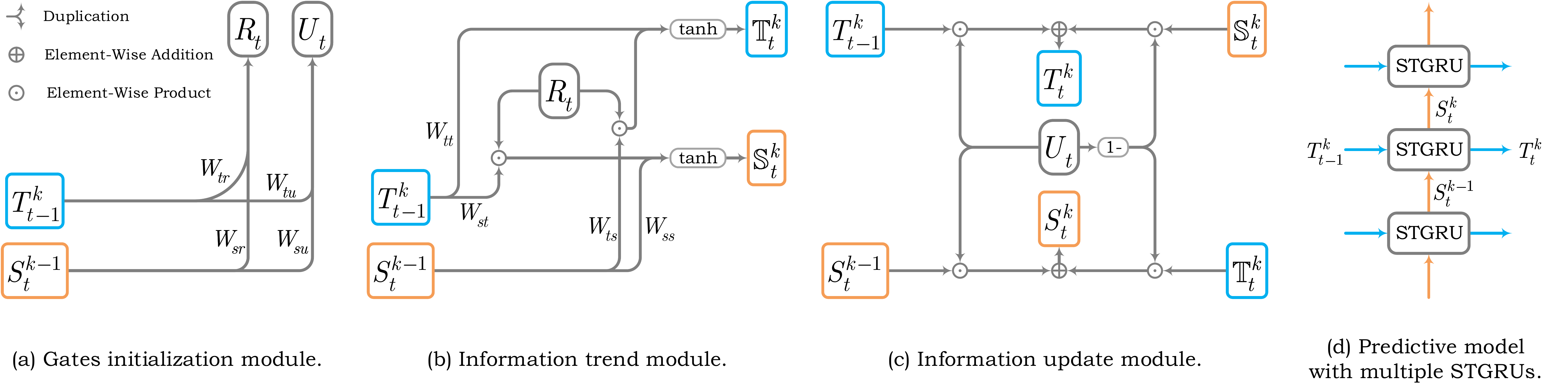}
  \caption{The detailed structure of the proposed spatiotemporal gated recurrent unit (STGRU).}\label{fig:stgru}
\end{figure*}

In addition, we design a new predictive memory (STGRU) to help preserve the spatiotemporal information during the state transitions using more efficient state transitions compared with current LSTM-based memories, as shown in Fig. \ref{fig:stgru}.
In the following sub-sections, we will show how the MGST-AE and STGRU can help preserve spatiotemporal information during the feature extraction and the state transitions, respectively.
%multiple encoders and decoders are utilized to process the videos in both temporal and spatial domains, respectively. In this way, the temporal and spatial information will be encoded and decoded independently and more spatiotemporal information can be preserved. On the other hand, multiple interactions between encoders and decoders have been employed using skip-connections, which can help the decoders recall multi-level visual information from the encoders. Overall, the predicting procedures are summarized as follows,
\subsubsection{Multi-Grained Spatiotemporal Auto-encoder (MGST-AE)}
Compared with the standard auto-encoder structure, the proposed MGST-AE mainly benefits from two schemes, the Spatiotemporal Encoding-Decoding (STED) Scheme and the Multi-Grained Information Recalling (MGIR) Scheme.

Compared with images, videos contain more complex spatiotemporal information. However, almost all predictive models will first encode the video inputs to low-dimensional features with merely a single 2D/3D encoder for each time step, after which the predictive memories have to further extract the temporal information and the spatial information from the encoded features for subsequent processing, which is information-lossy and greatly restricts the efficiency of the state transitions in the predictive memories. To solve this problem, we propose the spatiotemporal encoding-decoding (STED) scheme.
%lots of spatiotemporal information is abandoned, making it harder for decoders to correctly reconstruct them back to the high-dimensional data space. To solve this problem, we propose the spatiotemporal encoding scheme to

In our method, for each time step $t$, there is a video clip $v_{t-L+1:t}$ consisting of $L$ successive frames $\{v_{t-L+1},...,v_{t}\}$ as the input. Then the video clip $v_{t-L+1:t}$ is encoded by both the temporal and spatial encoders instead of a single encoder, respectively. In this way, the temporal and spatial information of the video inputs can be independently encoded in the temporal and spatial domains in advance, which can help avoid the unnecessary interactions between temporal and spatial information and greatly improve the efficiency of the following predictive memories. In particular, the spatiotemporal encoders are implemented based 3D convolutional networks due to their great power in extracting spatiotemporal features for videos \cite{ji20123d} and the spatiotemporal encoding procedures can be represented as follows,
\begin{eqnarray}\label{equ:encoders}
  T_E &=& Enc_T(v_{t-L+1:t}), \nonumber\\
  S_E &=& Enc_S(v_{t-L+1:t}),
\end{eqnarray}
where $Enc_T(\cdot), Enc_S(\cdot)$ denote the temporal and spatial encoders, respectively. $T_E, S_E$ denote the encoded temporal and spatial features, respectively.
%\subsubsection{Spatiotemporal Feature Prediction}

To efficiently model the state transitions, a new spatiotemporal predictive memory is designed on the basis of the Gated Recurrent Unit (GRU) \cite{cho2014learning}, which is denoted as STGRU. STGRU is much computation-cheaper compared with the popular LSTM-based memories and can obtain more satisfactory performance. The detailed structure of STGRU will be described in Section \ref{sec:stgru}. With the help of STGRU, the spatiotemporal features of future frames can be predicted as follows,
\begin{equation}
  (T_P,S_P) = STGRU(T_E,S_E),
\end{equation}
where $T_P,S_P$ denote the predicted temporal and spatial features, respectively.
%\subsubsection{Recalling-based Decoding}

For most of the predictive models, decoders can only utilize the predicted features to reconstruct the future frames, which are far from enough to generate fine visual details. One main reason for this problem is that the predicted deep features for the decoders only contain the high-level semantic features. The low-level texture information has been selectively abandoned by the front layers of the encoders. To solve this problem, we design the multi-grained information recalling (MGIR) scheme, which novelly employs direct interactions between the corresponding layers of the encoders and decoders using multiple skip-connections, as shown in Fig. \ref{fig:system model}. In this way, multi-grained encoded information can be easily recalled back by the decoders to reconstruct much better visual details and the detailed process can be represented as follows,
\begin{eqnarray}\label{equ:recall}
  %TDF_t^{l=0} &=& TPF_t,\nonumber\\
  %  SDF_t^{l=0} &=& SPF_t,\nonumber\\
  {T_D}^l &=& {Dec_T}^l({T_D}^{l-1}+\lambda {T_E}^{-l}),\nonumber\\
  {S_D}^l &=& {Dec_S}^l({S_D}^{l-1}+\lambda {S_E}^{-l}),
\end{eqnarray}
where ${Dec_T}^l(\cdot),{Dec_S}^l(\cdot)$ denote the $l^{th}$ layer of the temporal and spatial decoders, respectively. ${T_D}^l, {S_D}^l$ denote the decoded features from the $l^{th}$ layer in temporal and spatial decoders, respectively. ${T_E}^{-l}, {S_E}^{-l}$ denote the encoded features from the $l^{th}$ from the last layer in temporal and spatial encoders, respectively. $\lambda$ denotes the weight of the recalled information from the encoders.

The predicted frame $\hat{v}_{t+1}$ at time step $t$ can be represented as follows,
\begin{equation}
  \hat{v}_{t+1}=W_{1\times1}\ast([T_D,S_D]),
\end{equation}
where $T_D,S_D$ denote the decoded features from the top layers of the temporal and spatial decoders, respectively. $W_{1\times1}$ denotes the convolutional layer with kernel size $1\times1$. $\ast, [\cdot]$ denote the convolutional and the channel concatenating operations, respectively. Using the proposed MGST-AE, the spatiotemporal information can be greatly preserved during the feature extraction.
\subsubsection{Spatiotemporal Gated Recurrent Unit (STGRU)}\label{sec:stgru}
In this section, we elaborate how the proposed STGRU can help preserve spatiotemporal information during the state transitions.

As the most important component of predictive models, predictive memories have been widely discussed in a variety of works. In particular, due to the great power in capturing long-short term dependencies, LSTM-based memories can be the most popular one and have been widely employed in most of the latest models for video prediction, such as PredRNN (NeurIPS2017) \cite{wang2017predrnn}, PredRNN++ (ICML2018) \cite{wang2018predrnn++}, E3D-LSTM (ICLR2019) \cite{wang2019eidetic}, CrevNet (ICLR2020) \cite{yu2020efficient} and so on. However, the temporal states and the spatial states in these memories are easily influenced by the hidden state and some useful spatiotemporal information may be inevitably abandoned. In addition, the state transitions in LSTM-based memories are typically complex, restricting the efficiency of the state transitions. Motivated by these problems, we design a new predictive memory named STGRU based on the computation-cheaper GRU structure. The proposed STGRU only focuses on the spatiotemporal states and discards the hidden state. As a result, the spatiotemporal information can be well-preserved during the state transitions and much computation load will be saved.

%compared with LSTMs, GRUs\cite{cho2014learning} are much computation-cheaper while similar performance can still be obtained. Motivated by this, we design a spatiotemporal gated recurrent unit (STGRU) for video prediction, which can achieve similar performance compared with most of the LSTM-based memories while the computation load is much lower, especially in predicting high-resolution videos.
%\begin{figure}[t]
%  \centering
%  \includegraphics[width=\columnwidth]{STIP.pdf}
%  \caption{The architecture of multi-layer STGRUs.}\label{fig:stip}
%\end{figure}
The proposed STGRU consists of three modules, as shown in Fig. \ref{fig:stgru}: the gates initialization module, the information trend module and the information update module.
Compare with LSTM-based memories, there are only two gates in each STGRU (ConvLSTM: four gates, PredRNN: seven gates, E3D-LSTM: eight gates, CrevNet: five gates), as shown in the gate initialization module in Fig.\ref{fig:stgru}(a). The whole process can be represented as follows,
\begin{eqnarray}
  R_t &=& \sigma(W_{sr}*S_t^{k-1}+W_{tr}*T_{t-1}^k), \nonumber\\
  U_t &=& \sigma(W_{su}*S_t^{k-1}+W_{tu}*T_{t-1}^k),
\end{eqnarray}
where $R_t$ denotes the trend gate, which is utilized to model the spatiotemporal trend information in videos. $U_t$ denotes the update gate, which is utilized to update current spatiotemporal information to the future spatiotemporal information. As shown in Fig.\ref{fig:stgru}(d), multiple STGRUs are typically stacked to predict more reliable deep spatiotemporal features, and $T_t^k, S_t^k$ denote the predicted temporal and spatial features from the $k^{th}$ STGRU at time step $t$.

Using the trend gate, the temporal and spatial trend information can be predicted in the information trend module,
\begin{eqnarray}
  \mathbb{T}_t^k &=& \tanh(W_{tt}*T_{t-1}^k+R_t\odot(W_{st}*S_t^{k-1})), \nonumber\\
  \mathbb{S}_t^k &=& \tanh(W_{ss}*S_t^{k-1}+R_t\odot(W_{ts}*T_{t-1}^k)),
\end{eqnarray}
where $\mathbb{T}_t^k, \mathbb{S}_t^k$ denote the learned temporal and spatial trend information, respectively.

Using the learned changing trend information, the update module will update the current spatiotemporal information to the future spatiotemporal information,
\begin{eqnarray}
  T_t^k &=& (1-U_t)\odot \mathbb{T}_t^k + U_t\odot T_{t-1}^k, \nonumber\\
  S_t^k &=& (1-U_t)\odot \mathbb{S}_t^k + U_t\odot S_t^{k-1}.
\end{eqnarray}
The predicted temporal and spatial features $T_t^k, S_t^k$ consist of two terms, where the first term denotes the predicted spatiotemporal trend information and the second term denotes the current spatiotemporal information.
%In particular, as shown in Fig. \ref{fig:stip}, the initialized temporal and spatial states
%for the first STGRU are defined as follows,
%\begin{equation}
%% \nonumber to remove numbering (before each equation)
%  T_t^{k=0}, S_t^{k=0} = W_{1\times1}\ast[T_{t-1}^{k=0},T_E], S_E.
%\end{equation}
%The final predicted spatiotemporal features from multi-layer STGRUs are represented as follows,
%\begin{equation}
%  T_t^{k=N}, S_t^{k=N} = TPF_t, SPF_t,
%\end{equation}
%where $N$ denotes the number of the employed STGRUs.

\begin{table*}[b]
  \centering
  \setlength\tabcolsep{5.5pt}
  \caption{The experimental settings for different datasets. \textbf{Train} and \textbf{Test} denote the temporal period of the inputs and predictions while training and testing. \textbf{Layers} denotes the number of the stacked predictive units. \textbf{Patch} denotes the parameter of the PixelUnshuffle layer in Fig. \ref{fig:system model}, which can squeeze the input from $C\times H\times W$ to $(C\times Patch^2)\times (H/Patch)\times (W/Patch)$. \textbf{Hidden} denotes the number of the channels of the hidden states in STGRU, encoders, decoders and the discriminator. \textbf{DL} and \textbf{UL} denote the downsampling layer and the upsampling layer in the encoders and decoders, respectively. In particular, $DL_D, FC$ denote the downsampling layer and the fully-connected layer in the discriminator. The parameters of $DL, UL, DL_D, FC$ are summarized in Table \ref{tab:parameter_setting}.}
  \label{tab:experimental_setting}
  {\begin{tabular}{lcccccccccccc}
  \toprule
  \multicolumn{12}{c}{Experimental Settings}\\
  Dataset&Resolution &Train &Test    &Layers &Patch&Encoder&Decoder&Discriminator&Hidden&$\lambda$ &$\gamma_1$ & $\gamma_2$\\
  \midrule
  UCF Sports \cite{rodriguez2008action}&$512\times512$    &$4\rightarrow1$ &$4\rightarrow6$&16&2&$4\times DL$&$4\times UL$&$8\times {DL}_D,FC$&64&0.1& 0.010      & 0.0010\\
  Human3.6M \cite{ionescu2013human3}&$1024\times1024$  &$4\rightarrow1$ &$4\rightarrow4$&16&4&$4\times DL$&$4\times UL$&$8\times {DL}_D,FC$&64&0.1& 0.010      & 0.0010\\
  SJTU4K \cite{song2013sjtu}&$2160\times3840$  &$4\rightarrow1$ &$4\rightarrow4$&16&8&$4\times DL$&$4\times UL$&$8\times {DL}_D,FC$&64&0.1& 0.005      & 0.0005\\
  %Human3.6M \cite{ionescu2013human3}                                               &$128\times128\times3$  &$4\rightarrow4$ &$4\rightarrow4$&16&64&1.0\\
  %TownCentreXVID \cite{benfold2011stable}                                           &$1088\times1920\times3$&$4\rightarrow1$ &$4\rightarrow4$&16&$4\times DSL$&$4\times USL$&64&1.0\\
  %Something-Somethingv2 \cite{goyal2017something}                                   &$128\times128\times3$&$(5,10)\rightarrow(15,10)$ &$(5,10)\rightarrow(15,10)$&16&$3\times DSL$&$3\times USL$&64&1.0\\
  \bottomrule
  \end{tabular}}
\end{table*}

\begin{figure}[b]
  \centering
  \includegraphics[width=\columnwidth]{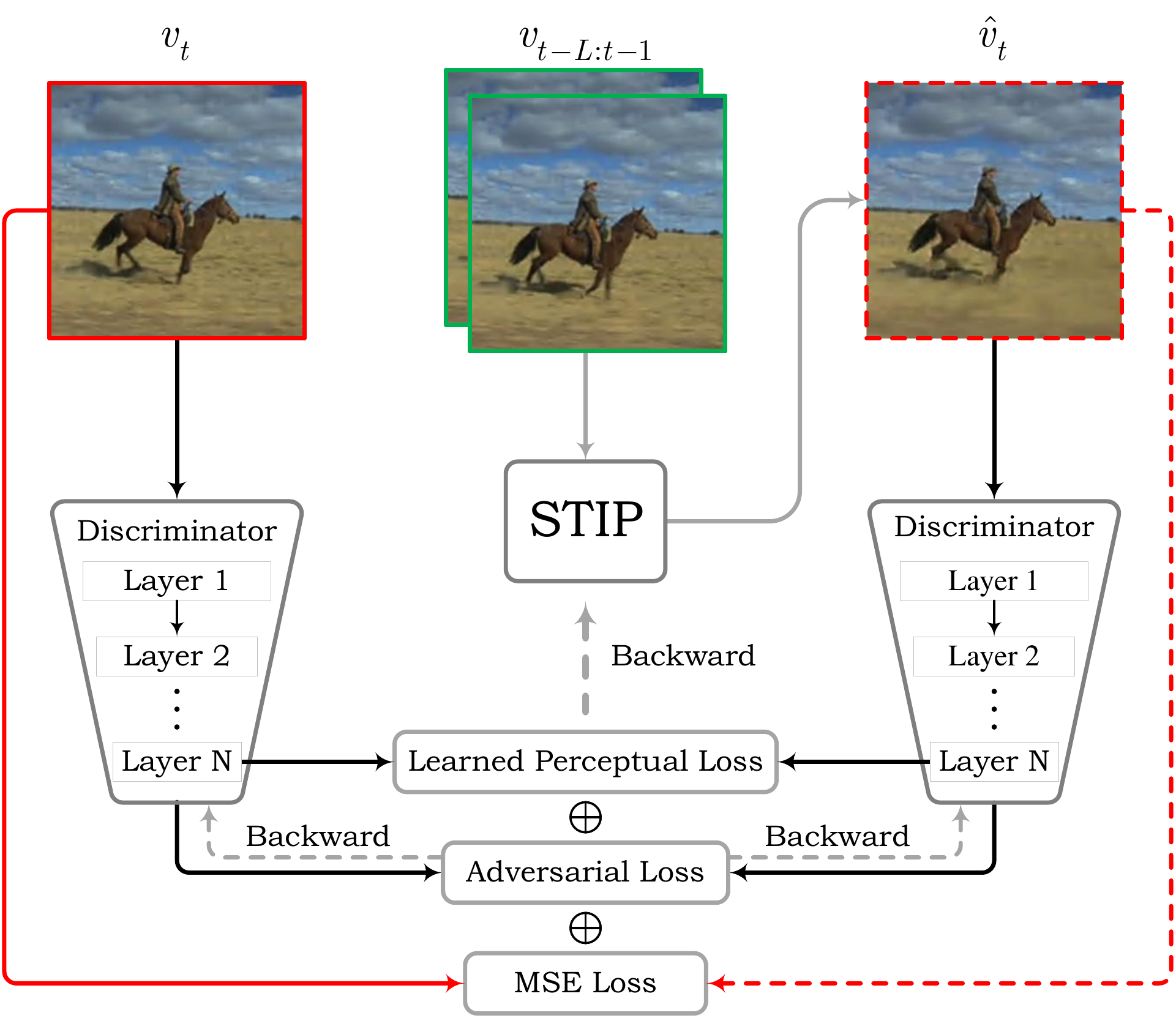}
  \caption{Training process. The MSE loss aims to optimize the objective quality, the adversarial loss aims to optimize the perceptual quality and the learned feature loss aims to balance these two terms.}\label{fig:loss}
\end{figure}

\subsection{Building a Perception-Augmented Predictive Model}
After solving the information loss problem, in this section, we aim to build a perception-augmented model by improving the perceptual-insensitive MSE-based loss functions. Recent works have begun to utilize GANs to help train the predictive models due to their advantages in generating naturalistic results. However, the MSE-based loss functions only focus on the objective quality and the adversarial loss only aims to improve the perceptual quality, making the training process extremely unstable, and it is difficult to achieve a satisfactory trade-off between both qualities. To solve this problem, an additional constraint urgently needs to be designed to balance these two terms.

Motivated by the advantages of the discriminators in GANs in learning the distributions of the video inputs, we propose to make use of the feature maps from the discriminators, which contain both the low-level objective features and the high-level perceptual features. 
And we design a new learned perceptual loss (LP-loss), as shown in the followings,
\begin{equation}
  \mathcal{L}_{LP} = \sum_{t=2}^{T}\mathcal{L}_2[D_l(v_t),D_l(\hat{v}_{t})],
\end{equation}
where $T$ denotes the total number of the time steps, $\mathcal{L}_2(\cdot)$ denotes the MSE loss function, $D_l(\cdot)$ denotes the $l^{th}$ layer of the discriminators (the feature map from the last layer is selected in our method). With the help of LP-loss, the perceptual quality and the objective quality can be tightly correlated, and thus the model is easier to train, and a more satisfactory trade-off between the objective and perceptual quality can be achieved.

\begin{algorithm}[!htb]
  \caption{Training the proposed model.}
  \label{alg1}
  \textbf{Input}: $V:\{v_1,...v_t...,v_{T-1}\}$\\
  \textbf{Parameter}: $W_{P}$, $W_{D}$\\
  \textbf{Output}: $\hat{V}:\{\hat{v}_2,...\hat{v}_t...,\hat{v}_{T}\}$
  \begin{algorithmic}[1] %[1] enables line numbers
    \REPEAT
    \STATE $V:$ random mini-batch from dataset
    \STATE Let time step $t=1$, layer $k=1$
    \STATE Initializing $T_{t=0}^k, k={1\sim N}$ for STGRUs
    \WHILE{$t<T$}
    \STATE $k=1$
    \STATE $(T_E, S_E) = (Enc_T(v_{t-L+1:t}), Enc_S(v_{t-L+1:t}))$
    \STATE $(T_{t-1}^k, S_t^{k-1})=(W_{1\times1}\ast[T_E,T_{t-1}^k], S_E)$
    \WHILE{$k\leq N$}
    \STATE $T_t^k, S_t^k= STGRU_k(T_{t-1}^k, S_t^{k-1})$
    \STATE $k=k+1$
    \ENDWHILE
    \STATE $l=1$
    \STATE $({T_D}^{l=0}, {S_D}^{l=0})=(T_t^{k=N}, S_t^{k=N})$
    \WHILE{$l\leq N_{layers}$}
    \STATE ${T_D}^l = {Dec_T}^l({T_D}^{l-1}+\lambda {T_E}^{-l})$
    \STATE ${S_D}^l = {Dec_S}^l({S_D}^{l-1}+\lambda {S_E}^{-l})$
    \STATE $l=l+1$
    \ENDWHILE
    \STATE $\hat{v}_t=W_{1\times1}([{T_D}^{l=N_{layers}}, {S_D}^{l=N_{layers}}])$
    \STATE $t=t+1$
    \ENDWHILE
    \STATE $W_{D}\stackrel{+}{\longleftarrow}-\nabla_{W_{D}}\mathcal{L}_{GAN}(D)$
    \STATE $W_{P}\stackrel{+}{\longleftarrow}-\nabla_{W_{P}}(\mathcal{L}_{MSE} + \gamma_1\mathcal{L}_{LP} + \gamma_2\mathcal{L}_{GAN}(P))$
    \UNTIL{convergence}
    \STATE \textbf{return} $\hat{V}:\{\hat{v}_2,...\hat{v}_t...,\hat{v}_{T}\}$
  \end{algorithmic}
\end{algorithm}
The whole training loss can be represented as follows,
\begin{equation}\label{eq:loss}
  \mathcal{L} = \mathcal{L}_{MSE}+\gamma_1\mathcal{L}_{GAN}(P) + \gamma_2{\mathcal{L}_{LP}},
\end{equation}
where $\gamma_1, \gamma_2$ denote the weights of the adversarial loss and the learned feature loss. In particular, $\mathcal{L}_{MSE}$ aims to optimize the objective quality, $\mathcal{L}_{GAN}(P)$ aims to optimize the perceptual quality and $\mathcal{L}_{LP}$ balances these two terms. The MSE loss and the adversarial loss are expressed as follows,
\begin{eqnarray}
  \mathcal{L}_{MSE} &=& \sum_{t=2}^{T}[\mathcal{L}_2(v_t,\hat{v}_t)],\nonumber\\
  \mathcal{L}_{GAN}(P) &=& -\sum_{t=2}^{T}[\log(D(\hat{v}_t))],\\
  \mathcal{L}_{GAN}(D) &=& -\sum_{t=2}^{T}[\log(D(v_t)) + \log(1-D(\hat{v}_t))],\nonumber
\end{eqnarray}
where $\mathcal{L}_{GAN}(D)$ denotes the adversarial loss for the discriminators in GANs and $\mathcal{L}_{GAN}(P)$ denotes the adversarial loss for the proposed STIP.
The detailed training procedures are summarized in Fig. \ref{fig:loss} and Alg. \ref{alg1}.

\begin{figure}[!htb]
  \centering
  \includegraphics[width=\columnwidth]{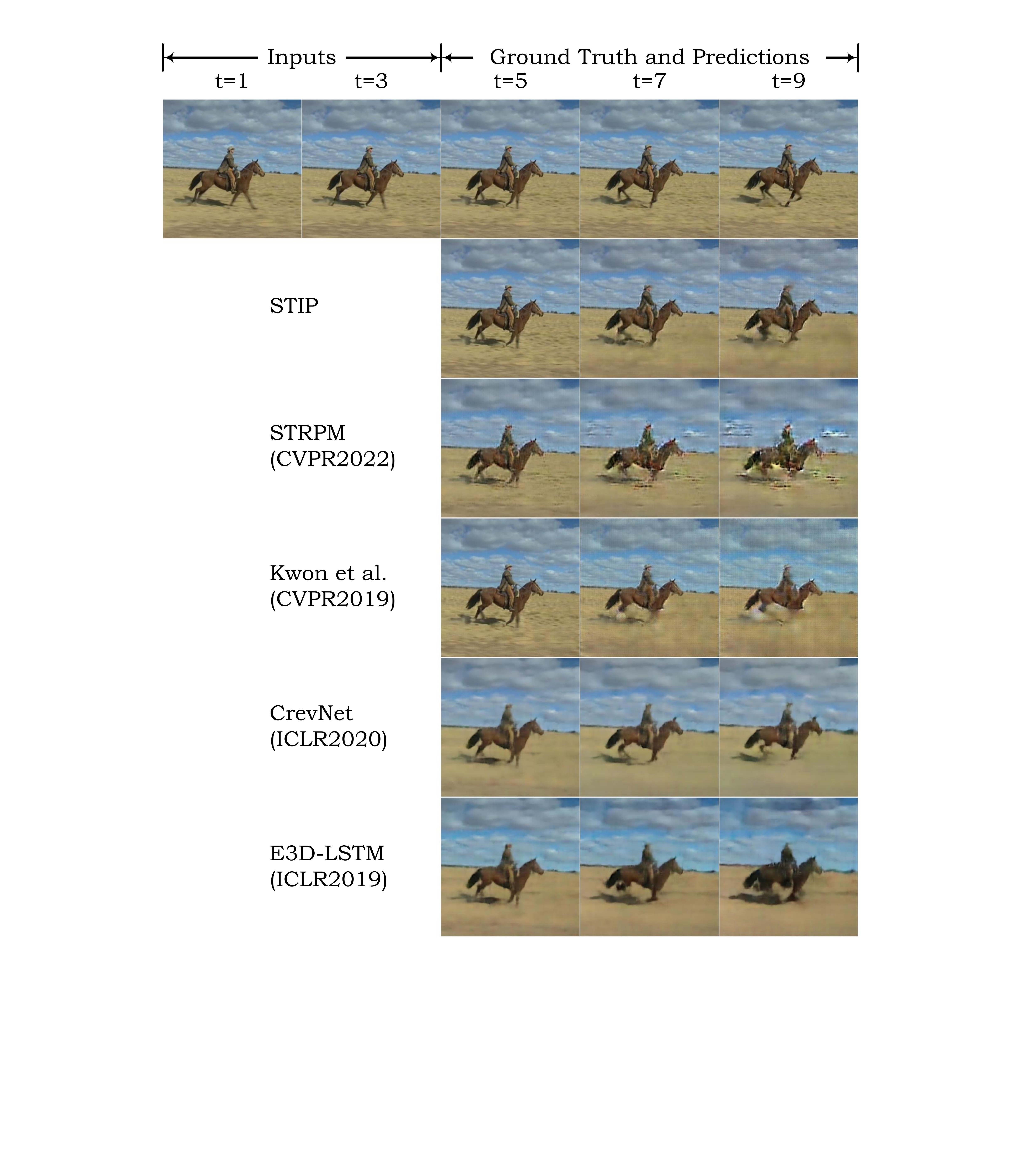}
  \caption{The generated examples on the UCF Sports dataset from different methods.}\label{fig:ucfsport}
  % \vspace{-0.2cm}
\end{figure}
\begin{figure*}[htb]
  \centering  
  \includegraphics[width=\textwidth]{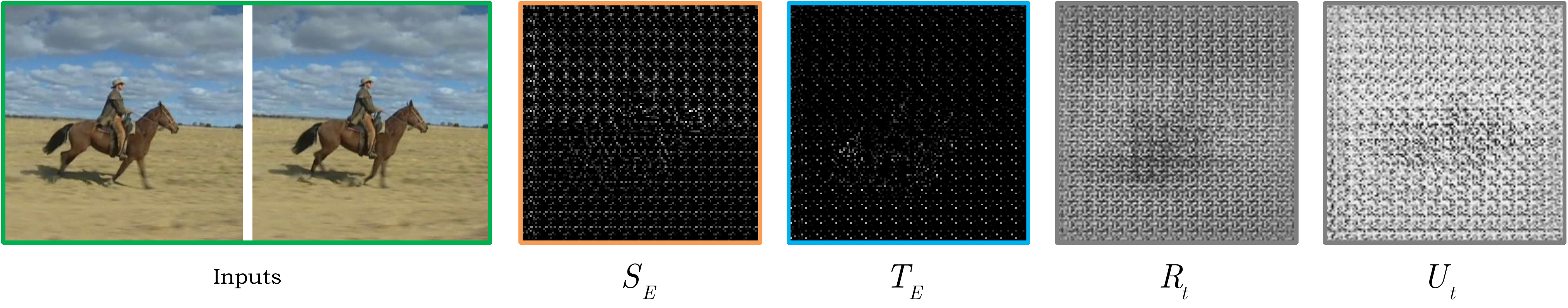}
  \caption{The visualized results of different features and gates in STIP. In particular, $S_E$, $T_E$ denote the encoded spatial and temporal features in Equation \ref{equ:encoders}. $R_t$ denotes the trend gate in STGRU, which aims to model the spatiotemporal changing trend information. $U_t$ denotes the update gate in STGRU, which is utilized to update current states to the predicted states using the learned trend information.}\label{fig:features}
\end{figure*}
\begin{table*}[htb]
  \centering
  \caption{The parameter settings for different layers. The encoders are built with the downsampling layers, which are utilized to extract deep features from video frames. The decoders are built with the upsampling layers, which are utilized to transform the features to the predicted frames. The discriminator is built with the downsampling layers and the fully-connected layer. The hidden layer denotes the integrated convolutional layers in STGRU.}
  \label{tab:parameter_setting}
  {\begin{tabular}{lccccccc}
  \toprule
  \multicolumn{8}{c}{Parameter Settings}\\
  Layers                                 &operation &Kernel &Stride    &Padding &Features&Normalization&Nolinearity\\
  \midrule
  Downsampling Layer (Encoder)                                   &convolution&3&2&1&64&-&LeakeyRelu(0.2)\\
  Downsampling Layer (Discriminator)                            &convolution&3&2&1&64&Group Normalization(4)&LeakeyRelu(0.2)\\
  Upsampling Layer (Decoder)                                    &deconvolution&3&2&1&64&-&LeakeyRelu(0.2)\\
  Hidden Layer (Unit)                                           &convolution&5&1&1&64&Layer Normalization&-\\
  Fully-Conected Layer (Discriminator)&fully-connected&-&-&-&64&-&Sigmoid()\\
  \bottomrule
  \end{tabular}}
\end{table*}

\section{Experimental Results}\label{sec:experiment}
\subsection{Experimental settings}
In this section, we evaluate the proposed model on 3 high-resolution datasets with resolutions ranging from 512 to 4K, and the experimental settings for each dataset are summarized in Table \ref{tab:experimental_setting}. 
%A total of 16 STGRUs are stacked to form the core predictive module in STIP. The kernel size of the convolutional layers in STGRU is $5\times 5$ with stride 1 for each dimension. 

The proposed model is implemented using PyTorch and optimized with Adam optimizer \cite{kingma2015adam}. The source code has been made public at \url{https://github.com/ZhengChang467/STIPHR}. The temporal length for each video clip input is set to 2. Layer normalization operations \cite{ba2016layer} are utilized to stabilize the training process for RNN models. We employ Mean Square Error (MSE), Peak Signal to Noise Ratio (PSNR), Structural Similarity (SSIM) \cite{wang2004image} scores to represent the objective quality and the Learned Perceptual Image Patch Similarity (LPIPS) \cite{zhang2018unreasonable} score to indicate the perceptual quality of the predictions.

\subsection{UCF Sports dataset ($512\times 512$)}

The UCF Sports dataset \cite{rodriguez2008action} consists of a set of actions collected from various sports, which are typically featured on broadcast television channels such as the BBC, SPN and so on. The resolution of this dataset is $480\times 720$. We sequentially sample 10 frames from the dataset, where 6,288 sequences are for training and 752 for testing. Each frame is resized to $512\times512$. All models are trained to predict the next frame and tested to predict the next 6 frames with the first 4 frames as the input.

\begin{table}[htb]
  \centering
  \setlength{\tabcolsep}{1.8mm}
  \caption{Quantitative results of different methods on the UCF Sports dataset (4 frames $ \rightarrow$ 6 frames).}\label{tab:ucfsport}
  { \begin{tabular}{lcccc}
      \toprule
      \multirow{2}{*}{Method}
                                                         & \multicolumn{2}{c}{$t=5$} & \multicolumn{2}{c}{$t=10$}\cr
      \cmidrule(lr){2-3} \cmidrule(lr){4-5}
                                                         & PSNR$\uparrow$            & LPIPS$\downarrow$             & PSNR$\uparrow$ & LPIPS$\downarrow$\cr
      \midrule
      ConvLSTM (NeurIPS2015) \cite{shi2015convolutional} & 26.43                     & 32.20                         & 17.80          & 58.78   \cr
      BeyondMSE (ICLR2016) \cite{mathieu2016deep}        & 26.42                     & 29.01                         & 18.46          & 55.28   \cr
      PredRNN (NeurIPS2017) \cite{wang2017predrnn}       & 27.17                     & 28.15                         & 19.65          & 55.34   \cr
      PredRNN++ (ICML2018) \cite{wang2018predrnn++}      & 27.26                     & 26.80                         & 19.67          & 56.79   \cr
      SAVP (arXiv 2018) \cite{lee2018stochastic}         & 27.35                     & 25.45                         & 19.90          & 49.91   \cr
      SV2P (ICLR2018) \cite{babaeizadeh2018stochastic}   & 27.44                     & 25.89                         & 19.97          & 51.33   \cr
      E3D-LSTM (ICLR2019) \cite{wang2019eidetic}         & 27.98                     & 25.13                         & 20.33          & 47.76   \cr
      CycleGAN (CVPR2019) \cite{kwon2019predicting}      & 27.99                     & 22.95                         & 19.99          & 44.93  \cr
      CrevNet (ICLR2020) \cite{yu2020efficient}          & 28.23                     & 23.87                         & 20.33          & 48.15   \cr
      MotionRNN (CVPR2021) \cite{wu2021motionrnn}        & 27.67                     & 24.23                         & 20.01          & 49.20                              \cr
      STRPM (CVPR2022) \cite{chang2022strpm}             & 28.54                     & 20.69                         & 20.59          & 41.11\cr
      \midrule
      STIP                                               & \textbf{30.75}            & \textbf{12.73}                & \textbf{21.83} & \textbf{39.67}    \cr
      \bottomrule
    \end{tabular}}
\end{table}

Fig. \ref{fig:ucfsport} shows the predicted examples from different methods, where STIP can predict more satisfactory results at all time steps, especially in the last 3 time steps. Table \ref{tab:ucfsport} shows the quantitative scores from different methods, where the proposed method outperforms others in all scores.

\begin{figure*}[htb]
  \centering  
  \includegraphics[width=\textwidth]{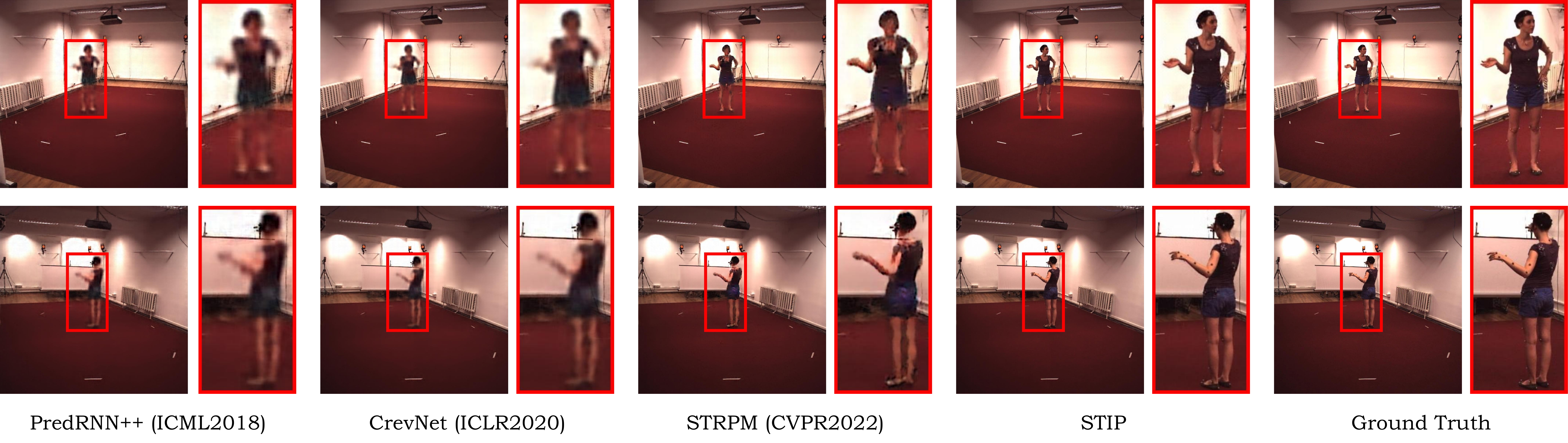}
  \caption{The generated examples on the Human3.6M dataset from different methods (4 frames $\longrightarrow$ 1 frame).}\label{fig:human36m}
\end{figure*}
Fig. \ref{fig:features} shows the visualized results of different features and gates in STIP. The input sample can be divided into two parts, the top part is the sky, which moves slowly but contains brighter pixels. The bottom part is the horse and the ground, which move fast but contain darker pixels. As shown in Fig. \ref{fig:features}, the encoded spatial features are corresponding to the pixel values in the samples, i.e. the top is brighter and the bottom is darker, which indicates that the spatial encoder can extract more appearance information from the inputs. However, the encoded temporal features pay more attention to the bottom part of the sample, i.e. the part with fast motions, which indicates the temporal encoder aims to extract features from motions. The visualized results of different features indicate that the proposed temporal and spatial encoders can extract different features from the temporal and spatial domains, respectively.
The visualized result of the update gate is brighter than the result of the trend gate, which indicates the trend gate aims to model the inter-frame residual information (a small part of the whole frame), but the update gate needs to reconstruct the redundant information between frames (a big part). 
% \begin{figure*}[t]
%   \centering
%   \includegraphics[width=\textwidth]{towncentre.pdf}
%   \caption{The generated examples on the TownCentreXVID dataset from different methods (4 frames $\longrightarrow$ 1 frame).}\label{fig:towncentre}
% \end{figure*}
\begin{figure*}[t]
  \centering
  \includegraphics[width=\textwidth]{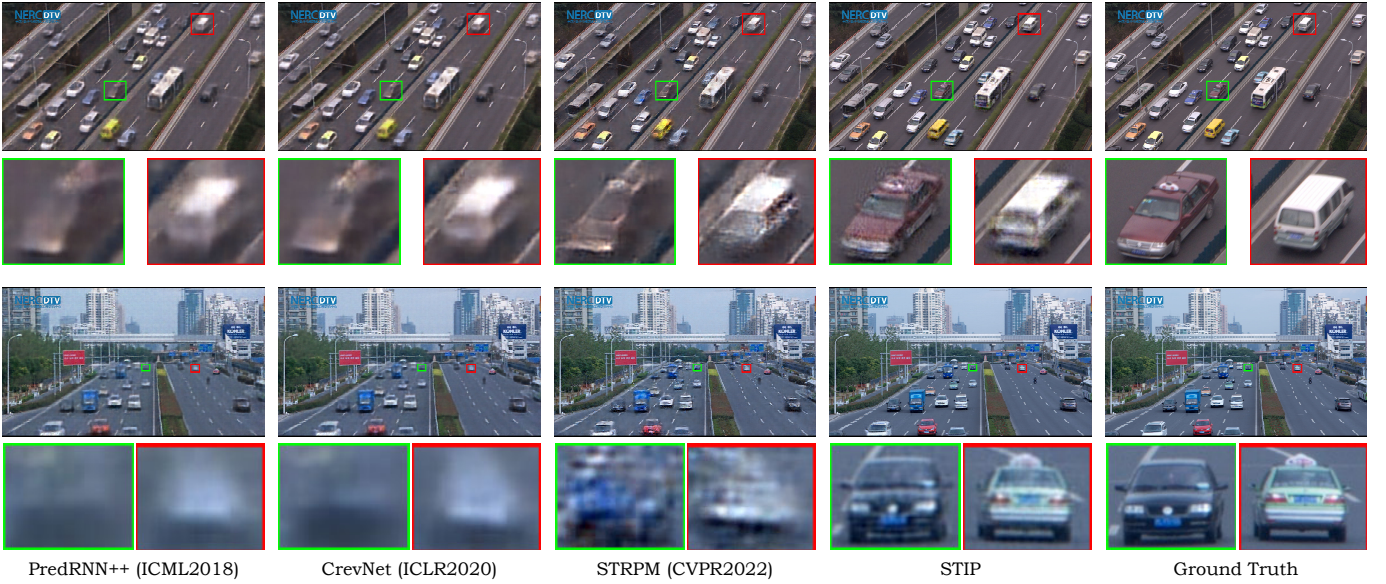}
  \caption{The generated examples on the SJTU4K dataset from different methods (4 frames $\longrightarrow$ 1 frame).}\label{fig:sjtu4k}
\end{figure*}
\begin{figure*}[t]
  \centering
  \includegraphics[width=\textwidth]{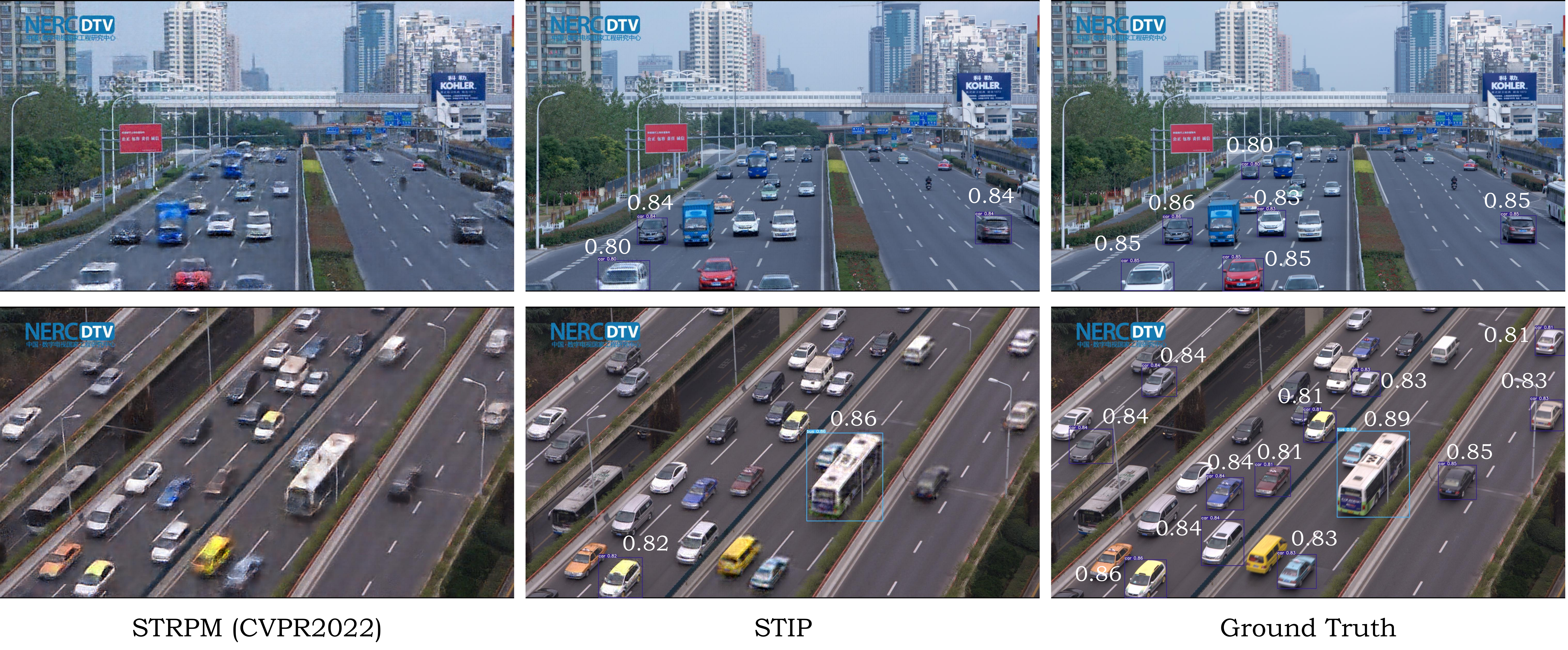}
  \caption{Object detection experiments on the generated examples from different methods on the basis of Yolov5s \cite{glenn_jocher_2021_4418161} pretrained model.}\label{fig:yolo}
\end{figure*}

\begin{table}[!htb]
  \centering
  %\fontsize{6.5}{8}\selectfont
  \setlength{\tabcolsep}{1.8mm}
  \caption{Quantitative results of different methods on the Human3.6M dataset (4 frames $ \rightarrow$ 4 frames).}
  \label{tab:human36m}
  { \begin{tabular}{lcccc}
      \toprule
      \multirow{2}{*}{Method}
      & \multicolumn{2}{c}{$t=5$} & \multicolumn{2}{c}{$t=8$}\cr
      \cmidrule(lr){2-3} \cmidrule(lr){4-5}
      & PSNR$\uparrow$            & LPIPS$\downarrow$            & PSNR$\uparrow$ & LPIPS$\downarrow$\cr
      \midrule
      ConvLSTM (NeurIPS2015) \cite{shi2015convolutional} & 29.88                     & 20.52                        & 26.25          & 22.31\cr
      PredRNN (NeurIPS2017) \cite{wang2017predrnn}       & 31.91                     & 12.62                        & 25.65          & 14.01\cr
      PredRNN++ (ICML2018) \cite{wang2018predrnn++}      & 32.05                     & 13.85                        & 27.51          & 14.94\cr
      SV2P (ICLR2018) \cite{babaeizadeh2018stochastic}   & 31.93                     & 13.91                        & 27.33          & 15.02\cr
      HFVP (NeurIPS2019) \cite{villegas2019high}         & 32.11                     & 13.41                        & 27.31          & 14.55\cr
      E3D-LSTM (ICLR2019) \cite{wang2019eidetic}         & 32.35                     & 13.12                        & 27.66          & 13.95\cr
      CycleGAN (CVPR2019) \cite{kwon2019predicting}      & 32.83                     & 10.18                        & 28.26          & 11.03\cr
      CrevNet (ICLR2020) \cite{yu2020efficient}          & 33.18                     & 11.54                        & 28.31          & 12.37\cr
      MotionRNN (CVPR2021) \cite{wu2021motionrnn}        & 32.20                     & 12.11                        & 28.03          & 13.29\cr
      STRPM (CVPR2022) \cite{chang2022strpm}             & 33.32                     & 9.74                         & 29.01          & 10.44 \cr
      \midrule
      STIP                                               & \textbf{36.82}            & \textbf{4.71}                & \textbf{30.99} & \textbf{10.35}  \cr
      \bottomrule
    \end{tabular}}
\end{table}
\subsection{Human3.6M dataset ($1024\times 1024$)}
The Human3.6M dataset \cite{ionescu2013human3} contains 3.6 million 3D human poses and the corresponding images conducted by 11 professional actors on 17 scenarios. Videos are captured by 4 calibrated cameras with a resolution of $1000\times1000$. Each frame is further resized to $1024\times1024$. A total of 73,404 sequences are for training and 8,582 sequences are for testing. All models are trained to predict the next frame and tested to predict the next 4 frames with the first 4 frames as the input.

Fig. \ref{fig:human36m} shows the qualitative results from different methods on the Human3.6M dataset, where the proposed method can predict more satisfactory appearance for the actor. In particular, compared with our previous work, more naturalistic results have been obtained. Table \ref{tab:human36m} summarizes the quantitative results, where the proposed STIP has achieved the best objective and subjective performance at all time steps.
%\begin{table*}[t]
%  \centering
%  %\fontsize{6.5}{8}\selectfont
%  \caption{Ablation studies of different predictive memories on the Moving MNIST dataset (10 frames $ \rightarrow$ 10 frames). The metrics are averaged over the predicted frames. \textbf{Parameters / Memory} denotes the number of parameters for each predictive memory.}
%  \label{tab:ablation_PM}
%   { \begin{tabular}{lccccc}
%    \toprule
%    Method&Backbone&MSE$\downarrow$&SSIM$\uparrow$&Parameters / Memory&Inference time / 800 samples\cr
%    \midrule
%    ConvLSTM (NeurIPS2015) \cite{shi2015convolutional}    &4$\times$ConvLSTMs&103.3 &0.707 &0.98M&16.47s                \cr
%    ST-LSTM (NeurIPS2017) \cite{wang2017predrnn}          &4$\times$ST-LSTMs&56.8  &0.869 & 1.57M &17.74s               \cr
%    Casual-LSTM (ICML2018) \cite{wang2018predrnn++}       &4$\times$Casual-LSTMs&46.5  &0.898 &1.80M&21.25s               \cr
%    MIM (CVPR2019) \cite{wang2019memory}               &4$\times$MIMs&44.2  &0.910 &3.03M&45.13s               \cr
%    E3D-LSTM (ICLR2019) \cite{wang2019eidetic}            &4$\times$E3D-LSTMs&41.3  &0.910 &4.70M&57.21s                \cr
%    RPM (ICLR2020) \cite{yu2020efficient}                 &4$\times$RPMs&23.7  &0.934 &1.77M&18.01s                \cr
%    MotionGRU (CVPR2021) \cite{wu2021motionrnn}      &4$\times$MotionGRUs&19.6  &0.951 &1.16M&17.58s                \cr
%    \midrule
%    STGRU   &4$\times$STGRUs&\textbf{12.8} &\textbf{0.955}&\textbf{0.79M}&\textbf{12.03s}\cr
%    \bottomrule
%    \end{tabular}}
%\end{table*}

\subsection{SJTU4K dataset ($2160\times 3840$)}
In this part, we evaluate the proposed method on a high-resolution surveillance video dataset, the SJTU4K dataset \cite{song2013sjtu}, with resolution of $2160\times 3840$. The SJTU4K dataset consists of 15 ultra-high resolution 4K videos with a wide variety of contents. A total of 3,873 training sequences and 445 testing sequences have been sampled from the SJTU4K dataset. All models are trained to predict the next frame and tested to predict the next 4 frames with the first 4 frames as the input.

Fig. \ref{fig:sjtu4k} shows the generated examples on the SJTU4K dataset from different methods. The proposed method significantly outperforms others in predicting the appearance of the cars. 
\begin{table}[htb]
  \centering
  \caption{Quantitative results of different methods on the SJTU4K test set (4 frames $ \rightarrow$ 4 frames). The inference time over 10 samples have also been summarized.}
  \label{tab:sjtu4k}
  { \begin{tabular}{lccc}
      \toprule
      \multirow{2}{*}{Method}
      & $t=5$ & $t=8$  & Inference\cr
      \cmidrule(lr){2-2} \cmidrule(lr){3-3}
      & PSNR$\uparrow$ / LPIPS$\downarrow$ & PSNR$\uparrow$ / LPIPS$\downarrow$ & Time\cr
      \midrule
      ConvLSTM \cite{shi2015convolutional} & 22.74 / 67.81  & 17.91 / 86.84  & 39.38s\cr
      PredRNN \cite{wang2017predrnn}       & 23.25 / 66.60  & 18.20 / 87.04  & 40.06s\cr
      PredRNN++ \cite{wang2018predrnn++}   & 23.43 / 64.07  & 18.55 / 86.34  & 53.11s\cr
      SAVP \cite{lee2018stochastic}        & 23.41 / 61.44  & 18.63 / 80.45  & 100.23s\cr
      CrevNet \cite{yu2020efficient}       & 24.35 / 62.31  & 19.61 / 80.91  & 52.98s\cr
      MotionRNN \cite{wu2021motionrnn}     & 23.47 / 65.21  & 19.72 / 81.39  & 61.87s\cr
      % \midrule
      %STRPM w/o Multi-encoders\&decoders                      &24.11  &63.68  &65.47    \cr
      %STRPM                                                   &\textbf{24.46}  &64.14    &65.65  \cr
      %STRPM + $L_{GAN}(P)$                                      &24.30  &55.11    &58.76  \cr
      %STRPM + $L_{GAN}(P)$ + $L_{LP}$                             &\textbf{24.23}  &\textbf{47.02}    &\textbf{48.95}  \cr
      STRPM \cite{chang2022strpm}         & 24.37 / 57.12   & 19.77 / 66.68 & 39.84s  \cr
      \midrule
      STIP                                & \textbf{28.91} / \textbf{29.77}   & \textbf{23.88} / \textbf{47.56} &\textbf{30.02s}\cr
      \bottomrule
    \end{tabular}}
\end{table}
Overall, for videos with relatively high-resolutions (4K), current methods can not achieve satisfactory performance due to the unacceptable information loss, while this problem has been solved by the proposed model STIP, and a series of detailed ablation studies will be conducted in section \ref{sec:ablation}. In particular, Table \ref{tab:sjtu4k} shows the quantitative results of different methods, where the proposed method has obtained the best objective and perceptual scores.
% \begin{table}[htb]
%   \centering
%   \caption{Object detection experiments on the generated examples from different methods on the basis of Yolov5s \cite{glenn_jocher_2021_4418161} pretrained model.}\label{tab:yolo}
%   { \begin{tabular}{lcc}
%       \toprule
%       \multirow{2}{*}{Method}
%                                                          & TownCentreXVID             & SJTU4K\cr
%       \cmidrule(lr){2-2} \cmidrule(lr){3-3}
%                                                          & Detected Persons$\uparrow$ & Detected Cars$\uparrow$\cr
%       \midrule
%       ConvLSTM (NeurIPS2015) \cite{shi2015convolutional} & 3                          & 0    \cr
%       PredRNN (NeurIPS2017) \cite{wang2017predrnn}       & 3                          & 1     \cr
%       PredRNN++ (ICML2018) \cite{wang2018predrnn++}      & 4                          & 0     \cr
%       CrevNet (ICLR2020) \cite{yu2020efficient}          & 3                          & 3     \cr
%       \midrule
%       STIP                                               & \textbf{10}                & \textbf{11}     \cr
%       Ground Truth                                       & \textbf{10}                & \textbf{11}     \cr
%       \bottomrule
%     \end{tabular}}
% \end{table}

To further evaluate the performance of the proposed model in predicting high-resolution videos, a series of object detection experiments are conducted on the predicted results from different methods, as shown in Fig. \ref{fig:yolo}. The objection detection experiments are employed on the basis of the latest Yolov5s \cite{glenn_jocher_2021_4418161} pretrained model. The confidence threshold is set to 0.80. Besides, the number of the detected objects is utilized to indirectly indicate the perceptual quality of the predictions from different methods. 
% As shown in Table \ref{tab:yolo}, the Yolov5s model can detect the most objects on the predicted frame from the proposed model. 
As shown in Fig. \ref{fig:yolo}, the confidences of the detected objects are higher on the predicted results from the proposed method, indicating the visual quality of the predictions from STIP is better than others.

%\begin{table}[t]
%\setlength\tabcolsep{2pt}
%  \centering
%  \begin{threeparttable}
%  \caption{Ablation studies of different predictive memories on Human3.6M dataset.}\label{tab:ablation_PM}
%   { \begin{tabular}{lccc}
%    \toprule
%    \multirow{2}{*}{Method}&\multirow{2}{*}{PSNR$\uparrow$}&Memory$\downarrow$&Inference Time$\downarrow$\cr
%    &&(1 sample)&(50 samples)\cr
%    \midrule
%    ConvLSTM (NeurIPS2015 )\cite{shi2015convolutional}                    &29.88&7578MB  &89s    \cr
%    ST-LSTM (NeurIPS2017) \cite{wang2017predrnn}                          &31.91&12948MB  &171s  \cr
%    Causal LSTM (ICML2018) \cite{wang2018predrnn++}                    &32.05&13957MB  &228s   \cr
%    RPM (ICLR2020) \cite{yu2020efficient}                              &33.01&6031MB  &68s       \cr
%    \midrule
%    STGRU                                            &\textbf{33.41}&\textbf{1382MB}  &\textbf{11s}  \cr
%    \bottomrule
%    \end{tabular}}
%    \end{threeparttable}
%\end{table}

\subsection{Ablation study}\label{sec:ablation}
In this section, a series of ablation studies have been conducted to evaluate the novelty of the proposed model.
\subsubsection{Evaluating the efficiency of the proposed Multi-Grained Spatiotemporal Auto-encoder}
In our method, the proposed multi-grained spatiotemporal auto-encoder benefits from two schemes, the spatiotemporal encoding (STED) scheme and the multi-grained information recalling (MGIR) scheme.

\begin{table}[!htb]
  \centering
  \caption{Ablation study of the information-preserving scheme during the feature extraction on different datasets (4 frames $ \rightarrow$ 1 frame). All models are optimized using the MSE loss function. PSNR score is employed to indicate the visual quality of the predictions. 
  }\label{tab:ablation_information}
  { \begin{tabular}{lccc}
      \toprule
      Methods      & UCF Sports      & Human3.6M       & SJTU4K\cr
      \midrule
      1. w/o STED, MGIR & 28.46          & 33.41                    & 24.01\cr
      2. w STED         & 28.78          & 33.98                    & 26.51\cr
      3. w MGIR        & 29.54          & 35.11                    & 27.32\cr
      4. w STED, MGIR   & \textbf{31.29} & \textbf{36.96}  & \textbf{29.65}\cr
      \bottomrule
    \end{tabular}}
\end{table}
%The first one is that multiple encoders and decoders are utilized to extract features and reconstruct future frames in temporal and spatial domains, respectively. In this way, the temporal and spatial information can be processed independently. The second approach is that direct interactions have been employed between encoders and decoders using skip-connections, where the decoders can recall multi-level visual information to generated satisfactory visual details. Especially, the recalling scheme has helped a lot while predicting high-resolution videos (1080p, 4K).
To discuss the efficiency of both schemes, experimental results have been summarized in Table \ref{tab:ablation_information}. All models are trained and tested to predict the next frame with the first 4 frames as the input, optimized with the MSE loss function.
% \begin{table}[!htb]
%   \centering
%   \setlength{\tabcolsep}{1.3mm}
%   \caption{Ablation study of the information-preserving scheme during feature extraction on different datasets (4 frames $ \rightarrow$ 1 frame). PSNR score is employed to indicate the visual quality of the predictions.
%   }\label{tab:ablation_information}
%   { \begin{tabular}{lcccc}
%       \toprule
%       Datasets      & UCF Sport      & Human3.6M      & TownCentreXVID & SJTU4K\cr
%       \midrule
%       w/o STE, SGIR & 28.46          & 33.41          & 32.12          & 24.01\cr
%       w STE         & 28.78          & 33.98          & 32.50          & 26.51\cr
%       w SGIR        & 29.54          & 35.11          & 32.89          & 27.32\cr
%       w STE, SGIR   & \textbf{31.29} & \textbf{36.96} & \textbf{33.98} & \textbf{29.65}\cr
%       \bottomrule
%     \end{tabular}}
% \end{table}

As shown in Table \ref{tab:ablation_information}, both the STED scheme and the MGIR scheme can help to obtain better PSNR scores on all datasets, indicating more visual information has been preserved by both approaches. In particular, STIP with MGIR scheme (method 3) outperforms STIP with STED scheme (method 2), indicating the decoder needs more direct information from the encoders for more reliable prediction and the extension from our previous work is very necessary.
%\begin{table*}[htb]
%  \centering
%  \caption{Ablation study of the information-preserving scheme during feature extraction on different datasets.
%  }\label{tab:ablation_information}
%   { \begin{tabular}{lcccccccc}
%    \toprule
%    Datasets&\multicolumn{2}{c}{UCF Sport}&\multicolumn{2}{c}{Human3.6M}&\multicolumn{2}{c}{TownCentreXVID}&\multicolumn{2}{c}{SJTU4K}\cr
%    \cmidrule(lr){2-3} \cmidrule(lr){4-5} \cmidrule(lr){6-7}\cmidrule(lr){8-9}
%     \diagbox{MGST-AE}{PSNR$\uparrow$}{ST-Encoding}&w/o&w&w/o&w&w/o&w&w/o&w\cr
%    \midrule
%    w/o               &28.46  &28.78           &33.41  &33.98        &32.12&32.50            &24.01&26.51     \cr
%    w             &29.54  &\textbf{31.29}  &35.11  &\textbf{36.96}  &32.89&\textbf{33.98}  &27.32&\textbf{29.65}     \cr
%    \bottomrule
%    \end{tabular}}
%\end{table*}

\subsubsection{Evaluating the efficiency of the proposed STGRU}
In this part, we conduct a series of ablation studies to evaluate the efficiency of the proposed predictive memory, STGRU. We build a series of predictive models with the same structure except for the predictive memory. Seven predictive models with different predictive memories are trained and tested on the Human3.6M dataset. For a fair comparison, we employ traditional encoders and decoders for all models with the same parameters. All models are trained with the standard MSE loss function.

\begin{table}[htb]
    \centering
    \caption{Quantitative results of predictive methods with different predictive units 
    % Ablation studies on the proposed residual predictive memory and the spatiotemporal encoding-decoding scheme (\textbf{STED}) 
    on the Human3.6M dataset (4 frames $\rightarrow$ 4 frames). PSNR and LPIPS scores are averaged over all 4 predictions.
    %\textbf{T}, \textbf{S} denote the temporal module and the spatial module in RPM, respectively.
    For a fair comparison, the encoders and decoders for all models are with the same structure and the number of the hidden state channels for all the memories is set to 128 with a kernel size of 5. We stack 16 memories into each model. All models are trained with MSE loss functions. The parameters and floating point operations (FLOPs) are recorded over 1 sample within 1 unit.}
    \label{tab:ablation_PM}
     { \begin{tabular}{lcccc}
      \toprule
      %\multicolumn{4}{c}{Human3.6M}\cr
      Method
      &PSNR$\uparrow$&LPIPS$\downarrow$&Parameters&FLOPs\cr
      \midrule
      %ST-LSTM \cite{wang2017predrnn}                              &28.54  &13.35&100.50M&28.81G\cr
      ConvGRU\cite{ballas2016delving}                   &28.12  &19.87  &2.34M  &0.59G\\
      TrajGRU-L9 \cite{shi2017deep}                     &28.34  &19.12  &1.68M  &0.42G\\
      dGRU \cite{finn2016unsupervised}                  &28.55  &19.03  &4.69M  &1.17G\\
      Casual-LSTM \cite{wang2018predrnn++}              &29.61  &14.42  &7.20M  &1.80G\cr
      E3D-LSTM \cite{wang2019eidetic}                   &29.93  &13.23  &18.80M &4.70G\cr
      Reversible-PM \cite{yu2020efficient}              &30.13  &12.62  &7.08M  &1.77G\cr
      
      % RPM w/o residual                                            &30.11  &12.64&108.29M&30.55G\cr
      % RPM ($\theta=1, \tau=1$)                                                &30.14  &12.58&\textbf{45.95M}&\textbf{14.96G}\cr
      % RPM ($\theta=5, \tau=1$)                                                  &30.56  &11.98&77.09M&22.75G\cr
      % RPM ($\theta=1, \tau=5$)                                                   &30.32  &12.31&77.09M&22.75G\cr
      RPM \cite{chang2022strpm}                        &31.10  &11.89&6.72M&1.68G\cr
      \midrule
      % RPM + STED                                                 &\textbf{31.81}&\textbf{11.72}&109.13M&34.24G\\
      % STGRU &31.21&11.02&51.29M &20.20G\\
      STGRU &\textbf{31.21}&\textbf{11.02}&3.14M
      &0.79G\\
      \bottomrule
      \end{tabular}}
  \end{table}

Table \ref{tab:ablation_PM} shows the quantitative results of different memories. The proposed STGRU can outperform other popular memories in the PSNR score with fewer parameters and faster inference speed.

\subsubsection{Evaluating the efficiency of the learned perceptual loss (LP-loss)}
Similar to our previous work \cite{chang2022strpm}, we also evaluate the performance of STIPs with different loss functions on the UCF Sports dataset and the Human3.6M dataset. For a fair comparison, all STIPs are set with the same parameters. The results are summarized in Table \ref{tab:ablation_loss}.
% \begin{table}[!htb]
%   \setlength\tabcolsep{5pt}
%   \centering
%   \caption{Ablation studies of different loss functions. Caltech Pedestrian: 10 frames $\longrightarrow$ 1 frame. TownCentreXVID: 4 frames $\longrightarrow$ 1 frame.}\label{tab:ablation_loss}
%   { \begin{tabular}{lcccc}
%       \toprule
%       \multirow{2}{*}{Loss Functions}                           & \multicolumn{2}{c}{Caltech Pedestrian} & \multicolumn{2}{c}{TownCentreXVID}\cr
%       \cmidrule(lr){2-3}\cmidrule(lr){4-5}
%                                                                 & MSE$\downarrow$                        & LPIPS$\downarrow$                     & PSNR$\uparrow$ & LPIPS$\downarrow$\cr
%       \midrule
%       $\mathcal{L}_{MSE}$                                       & \textbf{0.97}                          & 5.04                                  & \textbf{33.86} & 7.31   \cr
%       $\mathcal{L}_{MSE}+\mathcal{L}_{GAN}$                     & 2.56                                   & 4.54                                  & 32.87          & 6.32  \cr
%       \midrule
%       $\mathcal{L}_{MSE}+\mathcal{L}_{GAN} + \mathcal{L}_{LFL}$ & 1.54                                   & \textbf{3.91}                         & 33.57          & \textbf{4.32}  \cr
%       \bottomrule
%     \end{tabular}}
% \end{table}

\begin{table}[t]
  \centering
  %\fontsize{6.5}{8}\selectfont
  % \setlength\tabcolsep{3pt}
  \caption{Ablation studies on STRPM with different loss functions. Performance scores are averaged over all predictions.}
  \label{tab:ablation_loss}
   { \begin{tabular}{lcccc}
    \toprule
    %\multicolumn{4}{c}{Human3.6M}\cr
    \multirow{3}{*}{Method}
    &\multicolumn{2}{c}{UCF sports}&\multicolumn{2}{c}{Human3.6M}\cr
    &\multicolumn{2}{c}{$4\rightarrow6$}&\multicolumn{2}{c}{$4\rightarrow4$}\cr
    \cmidrule(lr){2-3} \cmidrule(lr){4-5}
    &PSNR$\uparrow$&LPIPS$\downarrow$&PSNR$\uparrow$&LPIPS$\downarrow$\cr
    \midrule
    %CrevNet + (MSE)                                     &28.23  &23.87  &33.18  &11.54\cr
%    CrevNet + (MSE,$L_{GAN}$)                            &27.50 &23.11 &32.44 &10.47\cr
%    CrevNet + (MSE,$L_{GAN}$,$L_{LP}$)                   &27.98 &22.83 &32.80 &10.36\cr
    $\mathcal{L}_{MSE}$                                                   &\textbf{26.01}  &35.72 &\textbf{34.11}  &10.44\cr
    $\mathcal{L}_{MSE}$+$\mathcal{L}_{GAN}$                               &24.99  &30.13  &31.45  &8.76\cr
    $\mathcal{L}_{MSE}$+$\mathcal{L}_{GAN}$+$\mathcal{L}_{LP}$            &25.47  &\textbf{27.67}  &33.53  &\textbf{7.54}\cr
    \bottomrule
    \end{tabular}}
\end{table}
From the quantitative results, STIP with $\mathcal{L}_{GAN}$ has obtained a lower LPIPS score compared with STIP merely with $\mathcal{L}_{MSE}$, indicating the adversarial loss can help improve the perceptual quality of the predictions. However, STIP with $\mathcal{L}_{MSE}$ has obtained better MSE and PSNR scores compared with STIP trained with adversarial loss, indicating the MSE loss can help improve the objective quality of the predictions. To balance both qualities, the learned perceptual loss (LP-loss) is proposed to help achieve a satisfactory trade-off between the perceptual quality and objective quality, as shown in Table \ref{tab:ablation_loss}.

\section{Conclusion}\label{sec:conclusion}
In this paper, we proposed a Spatiotemporal Information-Preserving and Perception-Augmented model (STIP) for high-resolution video prediction. In our method, we made a series of contributions to solve the information loss problem and improve the perception-insensitive MSE-based loss functions. On the one hand, we designed a Multi-Grained Spatiotemporal Auto-Encoder (MGST-AE) to preserve spatiotemporal information during the feature extraction and proposed a Spatiotemporal Gated Recurrent Unit (STGRU) to preserve spatiotemporal information during the state transitions. On the other hand, we further designed a new Learned Perceptual Loss (LP-loss) to help obtain a satisfactory trade-off between the objective quality and the perceptual quality. Experimental results showed the proposed method can achieve better performance compared with a variety of state-of-the-art methods.

\section{Limitations and Future Works}
Although noticeable improvements have been achieved in this paper for high-resolution video prediction, there are still some limitations that need to be considered in the future. Firstly, the performance in predicting multi-frame videos with high-resolutions is still not satisfactory. Secondly, the spatiotemporal features extracted from the predictive units are highly-desired to be employed in other tasks, such as video classification, object detection, etc. Thirdly, the generalization ability of current predictive models needs to be improved to achieve satisfactory performance in normal real-world scenarios. Future works should carefully address the above challenges.
%two approaches were employed to preserve more visual information for future video prediction. On the one hand, multiple encoders and decoders were utilized to process the video inputs in temporal and spatial domains respectively. In this way, the temporal and spatial information can be preserved independently. On the other hand, direct interactions between encoders and decoders were built using multiple skip-connections, which can help decoders recall multi-level visual information from the encoders to predict more satisfactory results. In addition, a new predictive memory, STGRU, was proposed to further reduce the computation load for STIP. Moreover, to predict results with higher perceptual quality, adversarial loss and a novel learned perceptual loss have been employed to train the proposed model using generative adversarial networks. Diverse experiments have been conducted to evaluate the proposed model. Experimental results showed that the proposed model can predict results with the highest perceptual and objective visual quality compared with various state-of-the-art methods.

%\appendices
%% you can choose not to have a title for an appendix
%% if you want by leaving the argument blank
%\section{}
%% use section* for acknowledgment
%\section*{Acknowledgment}

\ifCLASSOPTIONcaptionsoff
  \newpage
\fi



%\bibliographystyle{IEEEtran}
\bibliographystyle{IEEEtran}
\bibliography{STIP}

\begin{thebibliography}{10}
\providecommand{\url}[1]{#1}
\csname url@samestyle\endcsname
\providecommand{\newblock}{\relax}
\providecommand{\bibinfo}[2]{#2}
\providecommand{\BIBentrySTDinterwordspacing}{\spaceskip=0pt\relax}
\providecommand{\BIBentryALTinterwordstretchfactor}{4}
\providecommand{\BIBentryALTinterwordspacing}{\spaceskip=\fontdimen2\font plus
\BIBentryALTinterwordstretchfactor\fontdimen3\font minus
  \fontdimen4\font\relax}
\providecommand{\BIBforeignlanguage}[2]{{%
\expandafter\ifx\csname l@#1\endcsname\relax
\typeout{** WARNING: IEEEtran.bst: No hyphenation pattern has been}%
\typeout{** loaded for the language `#1'. Using the pattern for}%
\typeout{** the default language instead.}%
\else
\language=\csname l@#1\endcsname
\fi
#2}}
\providecommand{\BIBdecl}{\relax}
\BIBdecl

\bibitem{kang2014depth}
M.-K. Kang and K.-J. Yoon, ``Depth-discrepancy-compensated inter-prediction
  with adaptive segment management for multiview depth video coding,''
  \emph{IEEE Trans. Multimedia}, vol.~16, no.~6, pp. 1563--1573, 2014.

\bibitem{zupancic2016inter}
I.~Zupancic, S.~G. Blasi, E.~Peixoto, and E.~Izquierdo, ``Inter-prediction
  optimizations for video coding using adaptive coding unit visiting order,''
  \emph{IEEE Trans. Multimedia}, vol.~18, no.~9, pp. 1677--1690, 2016.

\bibitem{li2020spatio}
J.~Li, X.~Liu, W.~Zhang, M.~Zhang, J.~Song, and N.~Sebe, ``Spatio-temporal
  attention networks for action recognition and detection,'' \emph{IEEE Trans.
  Multimedia}, vol.~22, no.~11, pp. 2990--3001, 2020.

\bibitem{kalbkhani2016adaptive}
H.~Kalbkhani, M.~G. Shayesteh, and N.~Haghighat, ``Adaptive lstar model for
  long-range variable bit rate video traffic prediction,'' \emph{IEEE Trans.
  Multimedia}, vol.~19, no.~5, pp. 999--1014, 2016.

\bibitem{hu2020probabilistic}
A.~Hu, F.~Cotter, N.~Mohan, C.~Gurau, and A.~Kendall, ``Probabilistic future
  prediction for video scene understanding,'' in \emph{Eur. Conf. Comput.
  Vis.}, 2020, pp. 767--785.

\bibitem{kim2020dynamic}
N.-Y. Kim and J.~Kang, ``Dynamic motion estimation and evolution video
  prediction network,'' \emph{IEEE Trans. Multimedia}, 2020.

\bibitem{chen2019uni}
X.~Chen and W.~Wang, ``Uni-and-bi-directional video prediction via learning
  object-centric transformation,'' \emph{IEEE Trans. Multimedia}, vol.~22,
  no.~6, pp. 1591--1604, 2019.

\bibitem{liu2017video}
Z.~Liu, R.~A. Yeh, X.~Tang, Y.~Liu, and A.~Agarwala, ``Video frame synthesis
  using deep voxel flow,'' in \emph{Int. Conf. Comput. Vis.}, 2017, pp.
  4463--4471.

\bibitem{chang2022stam}
Z.~Chang, X.~Zhang, S.~Wang, S.~Ma, and W.~Gao, ``Stam: A spatiotemporal
  attention based memory for video prediction,'' \emph{IEEE Trans. Multimedia},
  2022.

\bibitem{ranzato2014video}
M.~Ranzato, A.~Szlam, J.~Bruna, M.~Mathieu, R.~Collobert, and S.~Chopra,
  ``Video (language) modeling: a baseline for generative models of natural
  videos,'' \emph{arXiv preprint arXiv:1412.6604}, 2014.

\bibitem{srivastava2015unsupervised}
N.~Srivastava, E.~Mansimov, and R.~Salakhudinov, ``Unsupervised learning of
  video representations using lstms,'' in \emph{Int. Conf. Mach. Learn.}, 2015,
  pp. 843--852.

\bibitem{shi2015convolutional}
X.~Shi, Z.~Chen, H.~Wang, D.~Y. Yeung, W.~K. Wong, and W.~Woo, ``Convolutional
  lstm network: A machine learning approach for precipitation nowcasting,'' in
  \emph{Adv. Neural Inform. Process. Syst.}, 2015.

\bibitem{finn2016unsupervised}
C.~Finn, I.~Goodfellow, and S.~Levine, ``Unsupervised learning for physical
  interaction through video prediction,'' \emph{Adv. Neural Inform. Process.
  Syst.}, vol.~29, 2016.

\bibitem{lotter2017deep}
W.~Lotter, G.~Kreiman, and D.~Cox, ``Deep predictive coding networks for video
  prediction and unsupervised learning,'' in \emph{Int. Conf. Learn.
  Represent.}, 2017.

\bibitem{villegas2017decomposing}
R.~Villegas, J.~Yang, S.~Hong, X.~Lin, and H.~Lee, ``Decomposing motion and
  content for natural video sequence prediction,'' in \emph{Int. Conf. Learn.
  Represent.}, 2017.

\bibitem{wang2017predrnn}
Y.~Wang, M.~Long, J.~Wang, Z.~Gao, and P.~S. Yu, ``Predrnn: Recurrent neural
  networks for predictive learning using spatiotemporal lstms,'' \emph{Adv.
  Neural Inform. Process. Syst.}, vol.~30, 2017.

\bibitem{oliu2018folded}
M.~Oliu, J.~Selva, and S.~Escalera, ``Folded recurrent neural networks for
  future video prediction,'' in \emph{Eur. Conf. Comput. Vis.}, 2018, pp.
  716--731.

\bibitem{wang2018predrnn++}
Y.~Wang, Z.~Gao, M.~Long, J.~Wang, and S.~Y. Philip, ``Predrnn++: Towards a
  resolution of the deep-in-time dilemma in spatiotemporal predictive
  learning,'' in \emph{Int. Conf. Mach. Learn.}, 2018, pp. 5123--5132.

\bibitem{wang2019eidetic}
Y.~Wang, L.~Jiang, M.-H. Yang, L.-J. Li, M.~Long, and L.~Fei-Fei, ``Eidetic 3d
  lstm: A model for video prediction and beyond,'' in \emph{Int. Conf. Learn.
  Represent.}, 2019.

\bibitem{wang2019memory}
Y.~Wang, J.~Zhang, H.~Zhu, M.~Long, J.~Wang, and P.~S. Yu, ``Memory in memory:
  A predictive neural network for learning higher-order non-stationarity from
  spatiotemporal dynamics,'' in \emph{IEEE Conf. Comput. Vis. Pattern Recog.},
  2019, pp. 9154--9162.

\bibitem{yu2020efficient}
W.~Yu, Y.~Lu, S.~Easterbrook, and S.~Fidler, ``Efficient and
  information-preserving future frame prediction and beyond,'' in \emph{Int.
  Conf. Learn. Represent.}, 2020.

\bibitem{lin2020self}
Z.~Lin, M.~Li, Z.~Zheng, Y.~Cheng, and C.~Yuan, ``Self-attention convlstm for
  spatiotemporal prediction,'' in \emph{AAAI}, vol.~34, no.~07, 2020, pp.
  11\,531--11\,538.

\bibitem{guen2020disentangling}
V.~L. Guen and N.~Thome, ``Disentangling physical dynamics from unknown factors
  for unsupervised video prediction,'' in \emph{IEEE Conf. Comput. Vis. Pattern
  Recog.}, 2020, pp. 11\,474--11\,484.

\bibitem{jin2020exploring}
B.~Jin, Y.~Hu, Q.~Tang, J.~Niu, Z.~Shi, Y.~Han, and X.~Li, ``Exploring
  spatial-temporal multi-frequency analysis for high-fidelity and
  temporal-consistency video prediction,'' in \emph{IEEE Conf. Comput. Vis.
  Pattern Recog.}, 2020, pp. 4554--4563.

\bibitem{wu2021motionrnn}
H.~Wu, Z.~Yao, J.~Wang, and M.~Long, ``Motionrnn: A flexible model for video
  prediction with spacetime-varying motions,'' in \emph{IEEE Conf. Comput. Vis.
  Pattern Recog.}, 2021, pp. 15\,435--15\,444.

\bibitem{lee2021video}
S.~Lee, H.~G. Kim, D.~H. Choi, H.-I. Kim, and Y.~M. Ro, ``Video prediction
  recalling long-term motion context via memory alignment learning,'' in
  \emph{IEEE Conf. Comput. Vis. Pattern Recog.}, 2021, pp. 3054--3063.

\bibitem{chang2022strpm}
Z.~Chang, X.~Zhang, S.~Wang, S.~Ma, and W.~Gao, ``Strpm: A spatiotemporal
  residual predictive model for high-resolution video prediction,'' in
  \emph{IEEE Conf. Comput. Vis. Pattern Recog.}, 2022.

\bibitem{hochreiter1997long}
S.~Hochreiter and J.~Schmidhuber, ``Long short-term memory,'' \emph{Neural
  Computation}, vol.~9, no.~8, pp. 1735--1780, 1997.

\bibitem{ballas2016delving}
N.~Ballas, L.~Yao, C.~Pal, and A.~C. Courville, ``Delving deeper into
  convolutional networks for learning video representations,'' in \emph{Int.
  Conf. Learn. Represent.}, 2016.

\bibitem{shi2017deep}
X.~Shi, Z.~Gao, L.~Lausen, H.~Wang, D.-Y. Yeung, W.-k. Wong, and W.-c. Woo,
  ``Deep learning for precipitation nowcasting: A benchmark and a new model,''
  \emph{Adv. Neural Inform. Process. Syst.}, vol.~30, 2017.

\bibitem{cho2014learning}
K.~Cho, B.~van Merrienboer, {\c{C}}.~G{\"u}l{\c{c}}ehre, D.~Bahdanau,
  F.~Bougares, H.~Schwenk, and Y.~Bengio, ``Learning phrase representations
  using rnn encoder-decoder for statistical machine translation,'' in
  \emph{EMNLP}, 2014.

\bibitem{wang2004image}
Z.~Wang, A.~C. Bovik, H.~R. Sheikh, and E.~P. Simoncelli, ``Image quality
  assessment: from error visibility to structural similarity,'' \emph{IEEE
  Trans. Image Process.}, vol.~13, no.~4, pp. 600--612, 2004.

\bibitem{mathieu2016deep}
M.~Mathieu, C.~Couprie, and Y.~LeCun, ``Deep multi-scale video prediction
  beyond mean square error,'' in \emph{Int. Conf. Learn. Represent.}, 2016.

\bibitem{kingma2013auto}
D.~P. Kingma and M.~Welling, ``Auto-encoding variational bayes,'' \emph{arXiv
  preprint arXiv:1312.6114}, 2013.

\bibitem{babaeizadeh2018stochastic}
M.~Babaeizadeh, C.~Finn, D.~Erhan, R.~H. Campbell, and S.~Levine, ``Stochastic
  variational video prediction,'' in \emph{Int. Conf. Learn. Represent.}, 2018.

\bibitem{denton2018stochastic}
E.~Denton and R.~Fergus, ``Stochastic video generation with a learned prior,''
  in \emph{Int. Conf. Mach. Learn.}, 2018, pp. 1174--1183.

\bibitem{franceschi2020stochastic}
J.-Y. Franceschi, E.~Delasalles, M.~Chen, S.~Lamprier, and P.~Gallinari,
  ``Stochastic latent residual video prediction,'' in \emph{Int. Conf. Mach.
  Learn.}, 2020, pp. 3233--3246.

\bibitem{xu2020video}
J.~Xu, H.~Xu, B.~Ni, X.~Yang, and T.~Darrell, ``Video prediction via example
  guidance,'' in \emph{Int. Conf. Mach. Learn.}, 2020, pp. 10\,628--10\,637.

\bibitem{wu2021greedy}
B.~Wu, S.~Nair, R.~Martin-Martin, L.~Fei-Fei, and C.~Finn, ``Greedy
  hierarchical variational autoencoders for large-scale video prediction,'' in
  \emph{IEEE Conf. Comput. Vis. Pattern Recog.}, 2021, pp. 2318--2328.

\bibitem{lee2018stochastic}
A.~X. Lee, R.~Zhang, F.~Ebert, P.~Abbeel, C.~Finn, and S.~Levine, ``Stochastic
  adversarial video prediction,'' \emph{arXiv preprint arXiv:1804.01523}, 2018.

\bibitem{kwon2019predicting}
Y.-H. Kwon and M.-G. Park, ``Predicting future frames using retrospective cycle
  gan,'' in \emph{IEEE Conf. Comput. Vis. Pattern Recog.}, 2019, pp.
  1811--1820.

\bibitem{chen2020long}
X.~Chen, C.~Xu, X.~Yang, and D.~Tao, ``Long-term video prediction via
  criticization and retrospection,'' \emph{IEEE Trans. Image Process.},
  vol.~29, pp. 7090--7103, 2020.

\bibitem{luc2020transformation}
P.~Luc, A.~Clark, S.~Dieleman, D.~d.~L. Casas, Y.~Doron, A.~Cassirer, and
  K.~Simonyan, ``Transformation-based adversarial video prediction on
  large-scale data,'' \emph{arXiv preprint arXiv:2003.04035}, 2020.

\bibitem{fujii2021x}
H.~Fujii, H.~Tanaka, M.~Ikeuchi, and K.~Hotta, ``X-net with different loss
  functions for cell image segmentation,'' in \emph{IEEE Conf. Comput. Vis.
  Pattern Recog.}, 2021, pp. 3793--3800.

\bibitem{schuldt2004recognizing}
C.~Schuldt, I.~Laptev, and B.~Caputo, ``Recognizing human actions: a local svm
  approach,'' in \emph{Int. Conf. Pattern Recog.}, vol.~3.\hskip 1em plus 0.5em
  minus 0.4em\relax IEEE, 2004, pp. 32--36.

\bibitem{babaeizadeh2017stochastic}
M.~Babaeizadeh, C.~Finn, D.~Erhan, R.~H. Campbell, and S.~Levine, ``Stochastic
  variational video prediction,'' in \emph{Int. Conf. Learn. Represent.}, 2018.

\bibitem{liang2017dual}
X.~Liang, L.~Lee, W.~Dai, and E.~P. Xing, ``Dual motion gan for future-flow
  embedded video prediction,'' in \emph{Int. Conf. Comput. Vis.}, 2017, pp.
  1744--1752.

\bibitem{ji20123d}
S.~Ji, W.~Xu, M.~Yang, and K.~Yu, ``3d convolutional neural networks for human
  action recognition,'' \emph{IEEE Trans. Pattern Anal. Mach. Intell.},
  vol.~35, no.~1, pp. 221--231, 2012.

\bibitem{rodriguez2008action}
M.~D. Rodriguez, J.~Ahmed, and M.~Shah, ``Action mach a spatio-temporal maximum
  average correlation height filter for action recognition,'' in \emph{IEEE
  Conf. Comput. Vis. Pattern Recog.}, 2008, pp. 1--8.

\bibitem{ionescu2013human3}
C.~Ionescu, D.~Papava, V.~Olaru, and C.~Sminchisescu, ``Human3. 6m: Large scale
  datasets and predictive methods for 3d human sensing in natural
  environments,'' \emph{IEEE Trans. Pattern Anal. Mach. Intell.}, vol.~36,
  no.~7, pp. 1325--1339, 2013.

\bibitem{song2013sjtu}
L.~Song, X.~Tang, W.~Zhang, X.~Yang, and P.~Xia, ``The sjtu 4k video sequence
  dataset,'' in \emph{International Workshop on Quality of Multimedia
  Experience}, 2013, pp. 34--35.

\bibitem{kingma2015adam}
D.~P. Kingma and J.~Ba, ``Adam: A method for stochastic optimization,'' in
  \emph{Int. Conf. Learn. Represent.}, 2015.

\bibitem{ba2016layer}
J.~L. Ba, J.~R. Kiros, and G.~E. Hinton, ``Layer normalization,'' \emph{STAT},
  vol. 1050, p.~21, 2016.

\bibitem{zhang2018unreasonable}
R.~Zhang, P.~Isola, A.~A. Efros, E.~Shechtman, and O.~Wang, ``The unreasonable
  effectiveness of deep features as a perceptual metric,'' in \emph{IEEE Conf.
  Comput. Vis. Pattern Recog.}, 2018, pp. 586--595.

\bibitem{glenn_jocher_2021_4418161}
\BIBentryALTinterwordspacing
G.~Jocher, A.~Stoken, J.~Borovec, NanoCode012, ChristopherSTAN, L.~Changyu,
  Laughing, tkianai, yxNONG, A.~Hogan, lorenzomammana, AlexWang1900,
  A.~Chaurasia, L.~Diaconu, Marc, wanghaoyang0106, ml5ah, Doug, Durgesh,
  F.~Ingham, Frederik, Guilhen, A.~Colmagro, H.~Ye, Jacobsolawetz,
  J.~Poznanski, J.~Fang, J.~Kim, K.~Doan, and L.~Yu, ``{ultralytics/yolov5:
  v4.0 - nn.SiLU() activations, Weights \& Biases logging, PyTorch Hub
  integration},'' Jan. 2021. [Online]. Available:
  \url{https://doi.org/10.5281/zenodo.4418161}
\BIBentrySTDinterwordspacing

\bibitem{villegas2019high}
R.~Villegas, A.~Pathak, H.~Kannan, D.~Erhan, Q.~V. Le, and H.~Lee, ``High
  fidelity video prediction with large stochastic recurrent neural networks,''
  \emph{Adv. Neural Inform. Process. Syst.}, vol.~32, 2019.

\end{thebibliography}


% \end{thebibliography}


% \end{IEEEbiography}


%\newpage



%\vfill


\end{document}